\documentclass{article} 
\usepackage{iclr2023_conference,times}
\usepackage{subfig}
\usepackage{bm}

\usepackage{amsmath,amsfonts,bm}

\def\eqref#1{equation~\ref{#1}}

\def\1{\bm{1}}

\DeclareMathAlphabet{\mathsfit}{\encodingdefault}{\sfdefault}{m}{sl}
\SetMathAlphabet{\mathsfit}{bold}{\encodingdefault}{\sfdefault}{bx}{n}

\newcommand{\E}{\mathbb{E}}

\newcommand{\Norm}{\text{N}} 
\newcommand{\obs}{\text{obs}} 
\newcommand{\mis}{\text{mis}} 

\newcommand{\bX}{\mathbf{X}}
\newcommand{\bx}{\mathbf{x}}
\newcommand{\bz}{\mathbf{z}}
\newcommand{\bZ}{\mathbf{Z}}
\newcommand{\btheta}{\bm{\theta}}
\newcommand{\bphi}{\bm{\phi}}
\newcommand{\bpsi}{\bm{\psi}}

\usepackage{hyperref}
\usepackage{url}
\usepackage{graphicx}
\graphicspath{ {./images/} }

\title{Leveraging Variational Autoencoders for Multiple Data Imputation}

\author{Breeshey Roskams-Hieter \& Jude Wells \\ 
HDR UK-Turing Wellcome\\
University of Edinburgh \& University College London\\
Edinburgh \& London, UK \\
\texttt{\{breeshey.roskams-hieter,jude.wells\}@hdruk.ac.uk} \\
\And
Sara Wade* \\
James Clerk Maxwell Building \\
University of Edinburgh \\
Edinburgh, EH9 3FD, UK \\
\texttt{\{sara.wade\}@ed.ac.uk} \\
}

\iclrfinalcopy 
\begin{document}

\maketitle

\begin{abstract}
Missing data persists as a major barrier to data analysis across numerous applications. Recently, deep generative models have been used for imputation of missing data, motivated by their ability to capture highly non-linear and complex relationships in the data. In this work, we investigate the ability of deep models, namely variational autoencoders (VAEs), to account for uncertainty in missing data through multiple imputation strategies. We find that VAEs provide poor empirical coverage of missing data, with underestimation and overconfident imputations, particularly for more extreme missing data values. To overcome this, we employ $\beta$-VAEs, which viewed from a generalized Bayes framework, provide robustness to model misspecification. Assigning a good value of $\beta$ is critical for uncertainty calibration and we demonstrate how this can be achieved using cross-validation. In downstream tasks, we show how multiple imputation with $\beta$-VAEs can avoid false discoveries that arise as artefacts of imputation.
\end{abstract}

\section{Introduction}

Missing data persists as a major barrier in large-scale analyses of multivariate data, due to issues like incomplete collection, data availability and low coverage. Early approaches for dealing with missing data tend to reduce the generalizability of results or skew the trends present in the data \citep{sinharay2001use}. These include listwise deletion, where only complete observations are considered, or imputation methods, where the complete data is used to predict plausible values for missing data points. Some of these imputation strategies include substitution by the mean of observed values, stochastic regression techniques and hot deck imputation. Single imputation approaches implicitly assume that the imputation is perfect and thereby fail to account for the uncertainty introduced by the prediction. An attractive solution for this is multiple imputation, which models the uncertainty in the missing values by producing several plausible values for each imputed data point. \citep{murray2018multiple}. The complete datasets are then analyzed in downstream tasks, and the results are combined to give estimates and standard
errors that acknowledge uncertainty in the missing data.

Recently, deep generative models have become an increasingly popular tool for imputing data, due to their ability to capture highly non-linear relationships and complex dependencies \citep[e.g.][]{qiu2020genomic, camino2019improving, gondara2018mida, lewis2021accurate,garcia2010pattern, nelwamondo2007missing, collier2020vaes, mattei2019miwae, ipsen2020not, nazabal2020handling, ma2018partial,ma2018eddi, ma2021identifiable,ma2020vaem}. 
For example, \citet{qiu2020genomic} use variational autoencoders (VAEs) for imputation of high-dimensional genomic data and find that it performs better than competing methods, such as singular value decomposition (SVD) and K-nearest neighbours (KNN), but focus solely on single imputation. In this work, we aim to investigate the ability of deep models, namely VAEs, to not only reconstruct the missing values but also account for uncertainty through multiple imputation strategies. 
While expressive and powerful, deep models have been shown to be overconfident even when predictions are incorrect \citep{szegedy2013intriguing} and also  underestimate the variability of out-of-sample test data \citep{nguyen2015deep}. In line with these results, we find that VAEs provide poor empirical coverage of the missing data, with underestimation and very overconfident imputations for more extreme missing data values that are far from the mean.

To overcome this, we employ $\beta$-VAEs \citep{higgins2017betavae}, which provide a framework for approximate Bayesian inference of deep generative models under the power likelihood. In statistics, inference based on the power likelihood has been shown to provide robustness against model misspecification \citep{bissiri2016general}, and thus, in our setting, it is crucial to avoid overfitting and achieve good coverage and well-calibrated uncertainty of the missing data.  As is well known in the literature, assigning a good value of $\beta$ is critical \citep{holmes2017assigning}, and we employ cross-validation to tune $\beta$ for accurate multiple imputation.  

Lastly, we study the implications of multiple imputation in downstream tasks. Failure to account for the uncertainty of the missing data in single imputation can lead to false confidence in downstream analyses, often yielding results that overestimate the significance of relationships between variables. In this paper, we demonstrate how using multiple imputation with $\beta$-VAEs yields fewer false positives and more acceptable false discovery rates. 

\section{Background}

\subsection{Variational autoencoders}

Variational autoencoders \citep{kingma2019introduction} combine graphical models and deep learning. They are made up of two parts, the \textbf{encoder} and \textbf{decoder}. Firstly, the encoder (also referred to as the inference model) takes an observed data point, \(\bx \in \mathbb{R}^D\), and computes the posterior distribution, \(p_{\btheta}(\bz|\bx)\), of the latent variables, $\bz \in \mathbb{R}^K$. As the true posterior is intractable in most cases, an approximate model, \( q_{\bphi}(\bz|\bx) \), is used to approximate the intractable true posterior, \( p_{\btheta}(\bz|\bx) \), and encode the  observed data into the latent variables. 
The second part is the decoder (also referred to as the generative model) where the latent variables, \(\bz\), are used to reconstruct data point, \(\hat{\bx}\), via the generative model, \( p_{\btheta}(\bx|\bz) \). The standard choice of distribution for both the inference and generative model is a simple, factorized Gaussian, where the Gaussian mean and variance are parametrized by neural networks, with $\bphi$ and $\btheta$ containing the weights and biases of the neural networks for the encoder and decoder,  respectively. Based on a training data set $\bX = (\bx_1, \ldots, \bx_N)$ containing $N$ data points, 
the neural network parameters \(\bphi\) and \(\btheta\) are optimized during training of the VAE by minimizing the reconstruction loss (i.e. minimizing the mean square error between reconstructed values, \(\hat{\bX} = (\hat{\bx}_1, \ldots, \hat{\bx}_N) \), and the observed values, \(\bX\)) and the latent loss (i.e. minimizing the Kullback-Leibler (KL) divergence between the variational posterior, \( q_{\bphi}(\bZ|\bX) \), and the standard Gaussian prior, \(p(\bZ)\), with $\bZ = (\bz_1,\ldots, \bz_N)$).

From a Bayesian perspective, this is equivalent to approximate variational inference of deep latent variable models, under the generative model $\bx_n \mid \bz_n \sim p_{\btheta} (\bx_n \mid \bz_n)$ 
with a Gaussian prior on latent variables $\bz_n \sim \Norm(\bm{0},\mathbf{I})$. To overcome intractability of the posterior, amortized variational inference \citep{gershman2014amortized} is employed,  assuming the variational posterior $q_{\bphi}(\bz_n|\bx_n)$ is parametrized by a neural network with $\bphi$ containing the weights and biases. The variational parameters $\bphi$ and generative model parameters $\btheta$ are optimized by minimizing the KL divergence between the variational posterior $q_{\bphi}(\bZ|\bX)$ and the true posterior $p_{\btheta}(\bZ|\bX)$, or equivalently 
maximizing the evidence lower bound (ELBO):
\begin{equation*} 
    \text{ELBO} = \sum_{n=1}^N\E_{\bz_n \sim q_{\bphi}(\bz_n|\bx_n)}[\log{p_{\btheta}(\bx_n|\bz_n)}] - D_{\text{KL}}(q_{\bphi}(\bz_n|\bx_n),p(\bz_n)).
\end{equation*}
During training, the ELBO is maximized by backpropagation through the hidden layers of the neural network, randomly sub-sampling the data at each training step and minimizing the loss through stochastic gradient descent. In order to compute the required gradients, we must employ the re-parameterization trick \citep{kingmaauto,rezende2014stochastic}, which uses a change of variables to obtain independence between the latent noise and $\bphi$.  

\subsubsection{$\beta$-VAEs}\label{sec:betaVAEs}

An extension on the classic VAE is the $\beta$-VAE, which includes a hyperparameter $\beta$ that enforces a regularization on the latent loss \citep{higgins2017betavae}. The optimization function is updated as follows:
\begin{equation}
    \text{ELBO} = \sum_{n=1}^N\E_{\bz_n \sim q_{\bphi}(\bz_n|\bx_n)}[\log{p_{\btheta}(\bx_n|\bz_n)}] - \beta \, D_{\text{KL}}(q_{\bphi}(\bz_n|\bx_n),p(\bz_n)). \label{eq:bvae}
\end{equation}
While in the machine learning community,  $\beta$-VAEs are motivated by their  improvement in disentangling the latent variables \citep{chen2018isolating}, we provide an alternative motivation from a statistical perspective. In particular, maximizing the $\beta$-VAE bound in \eqref{eq:bvae}, is equivalent to maximizing: 
\begin{equation*}
   \sum_{n=1}^N\E_{\bz_n \sim q_{\bphi}(\bz_n|\bx_n)}[\log{p_{\btheta}(\bx_n|\bz_n)^{1/\beta}}] -  \, D_{\text{KL}}(q_{\bphi}(\bz_n|\bx_n),p(\bz_n)), 
\end{equation*}
or minimizing the KL divergence between the variational posterior $q_{\bphi}(\bZ|\bX)$ and the posterior under the power likelihood (see Appendix \ref{app:bvae}): $$p_{\btheta,\beta}(\bZ|\bX) \propto \prod_{n=1}^N p_{\btheta}(\bx_n|\bz_n)^{1/\beta} p(\bz_n).$$
The use of the power likelihood in Bayesian statistics provides frequentist guarantees of posterior consistency in nonparametric models \citep{walker2001bayesian}, while the Bayesian model under the standard updating with $\beta=1$ may be inconsistent \citep{barron1999consistency}. Moreover, the power likelihood provides robustness to model misspecification \citep{bissiri2016general}. Given the complex, high-dimensional nature of the deep generative model $p_{\btheta}(\bx_n|\bz_n)$, this   acknowledges and allows for a mismatch between the generative model and the true data generating distribution.    

\subsection{Single imputation with VAEs}\label{sec:singimp_qiu}

Deep generative models have become an increasingly popular tool for imputation of missing data, due to their ability to accommodate highly non-linear relationships and complex dependencies in multivariate data. In this work, we focus on VAEs and, in particular, the approach of 
 \citet{qiu2020genomic}.  They first train the VAE using only the subset of complete data 
 to optimize the parameters \(\bphi\) and \(\btheta\). For each data point $n = 1,\ldots, N$, $\bx_n$ can be split into two parts:  $\bx_{\obs,n}$ containing the observed features and $\bx_{\mis,n}$ containing the missing features, where  $\bX_\obs = (\bx_{\obs,1},\ldots, \bx_{\obs,N}) $ and $\bX_\mis= (\bx_{\mis,1},\ldots, \bx_{\mis,N}) $ list the observed and missing data, respectively. For each data point with missing features, i.e. $\bx_n \neq \bx_{\obs,n}$, the optimal choice, under the squared error loss, is to impute with the mean under the generative model:
 \begin{align*}
    \widehat{\bx}_{\mis,n} &= \E \left[ \bx_{\mis,n} \mid \bx_{\obs,n} \right] 
    = \int \bx_{\mis,n} \, p_{\btheta}(\bx_{\mis,n} \mid \bx_{\obs,n}) \, d\bx_{\mis,n}\\
    &= \int \int \bx_{\mis,n} \, p_{\btheta}(\bx_{\mis,n}, \bz_n \mid \bx_{\obs,n})  \, d\bz_n d\bx_{\mis,n}.
\end{align*} 
This integral is intractable; thus in \citet{qiu2020genomic}, it is approximated by iteratively computing: 1) the expectation of $\bz_n$ (mean of the encoder) given $\widehat{\bx}_{\mis,n}$ and $\bx_{\obs,n}$:
     $$ \widehat{\bz}_n = \int \bz_n \, q_{\bphi}(\bz_n \mid \widehat{\bx}_{\mis,n}, \bx_{\obs,n}) \, d\bz_n, $$
    and 2) the expectation of $\widehat{\bx}_{\mis,n}$ (mean of the decoder) given $\widehat{\bz}_n$:
    $$ \widehat{\bx}_{\mis,n}  = \int \bx_{\mis,n} \, p_{\btheta}(\bx_{\mis,n} \mid \bx_{\obs,n},   \widehat{\bz}_{n}) \, d\bx_{\mis, n}. $$
The imputed values are initialized with zero imputation (i.e. mean imputation, after the initial standardization of the data), and these steps are repeated until convergence. Note that when the likelihood factorizes across features (e.g. factorized Gaussian), $p_{\btheta}(\bx_{\mis,n} \mid \bx_{\obs,n},  \bz_{n}) = p_{\btheta}(\bx_{\mis,n} \mid  \bz_{n})$.     
In their paper, \citet{qiu2020genomic} optimized the model and hyper-parameters through a grid search, claiming that the standard VAE ($\beta=1$) and training for 250 epochs resulted in the lowest mean absolute error of the imputed values when compared to true values.

\subsection{Multiple imputation}\label{sec:multipleimputation}

Multiple imputation \citep[see e.g.][]{little2019statistical, murray2018multiple, sinharay2001use} improves upon single imputation by retaining the mean and variance of the overall dataset and accounting for the uncertainty associated with the missing values. It does so by creating $M$ complete datasets with different plausible values for the missing data and then combining inference across all plausible datasets, e.g. by computing an overall mean and variance estimate for a certain statistic. The general standard is to create five to ten imputed datasets for optimal prediction, but more or less may be required depending on the fraction of data that is missing.

Multiple imputation leverages the missing at random (MAR) case, where variables in the observed data describe the missingness that is present in other variables. 
We aim to obtain and simulate from the predictive distribution for the missing data given the observed data, i.e. $p(\bX_\mis | \bX_\obs)$. In particular, we assume that $\bX$ follows a distribution, $p(\bX|\bpsi)$, where $\bpsi$ is a collection of all parameters of the model. Then we can write our predictive distribution as:
\begin{equation*}
    p(\bX_\mis|\bX_\obs) = \int p(\bX_\mis,\bpsi | \bX_\obs) d\bpsi 
    = \int p(\bX_\mis|\bX_\obs,\bpsi) p(\bpsi|\bX_\obs) d\bpsi.
\end{equation*}
To impute the missing data, and thereby simulate one of $M$ plausible datasets, data augmentation (DA) algorithms can be employed. Specifically, DA is a Markov chain method which iteratively samples 1) the parameters $\bpsi$ from the posterior $p(\bpsi|\bX_\obs, \bX_\mis)$ and 2) the missing data $\bX_\mis$ given $\bpsi$ from $p(\bX_\mis|\bX_\obs,\bpsi)$. This ultimately results in sampling from the predictive distribution  $p(\bX_\mis|\bX_\obs)$, producing one of the plausible datasets, denoted as $\bX_\mis^m = (\bx_{\mis,1}^m,\ldots,\bx_{\mis,N}^m)$. 
This procedure is repeated $M$ times to achieve $M$ plausible datasets. Inferences based on these $M$ imputed datasets can be combined via \textbf{Rubin's rules} to compute accurate inference about the entire dataset $\bX$. Note that statistical procedures must be done $M$ times, as there are $M$ datasets. 

\section{Methodology}\label{sec:methods}

In this work, we generalize single imputation with VAEs in two ways. First, we employ and compare three multiple imputation strategies to account for uncertainty in the missing data. Second, we extend using $\beta$-VAEs for improved robustness and uncertainty quantification. 

\subsection{Multiple imputation with $\beta$-VAEs}

In the case of multiple imputation, the latent variables, $\bZ$, of the $\beta$-VAE represent the parameters of our model, previously referred to as $\bpsi$ in Section \ref{sec:multipleimputation}. Therefore, to produce a sample from our target predictive distribution, $p_{\btheta,\beta}(\bX_\mis|\bX_\obs)$, we can iteratively sample from the joint distribution $p_{\btheta,\beta}(\bX_\mis, \bZ|\bX_\obs)$ via a Markov chain Monte Carlo scheme. For $\beta$-VAEs, the predictive distribution is constructed from the power likelihood, that is  the likelihood of our generative model is raised to the power 1/$\beta$ (see Section  \ref{sec:betaVAEs}):
\begin{equation}
    \label{eq:missdat}
    \begin{split}
    p_{\btheta,\beta}(\bX_\mis|\bX_\obs) &\propto \int p_{\btheta}(\bX_\mis|\bX_\obs,\bZ)^{1/\beta} p_{\btheta,\beta}(\bZ|\bX_\obs) d\bZ \\
    &=  \prod_{n=1}^N  \int p_{\btheta}(\bx_{\mis,n}|\bx_{\obs,n},\bz_n)^{1/\beta} p_{\btheta,\beta}(\bz_n|\bx_{\obs,n}) d\bz_n,
        \end{split}
\end{equation}
where standard VAEs correspond to $\beta=1$. We note that in the case of the factored Gaussian  generative model, the power likelihood  $p_{\btheta}(\bx_{\mis,n}|\bx_{\obs,n},\bz_n)^{1/\beta}$ is simply proportional to a Gaussian with variance rescaled by a factor of $\beta$. On the other hand, $p_{\btheta,\beta}(\bZ|\bX_\obs)$ represents the intractable true posterior of the latent variables under the power likelihood given the observed data only. 

In the following, we implement and compare three different approaches to sample from our target predictive distribution in \eqref{eq:missdat}: 1) pseudo-Gibbs (Section \ref{sec:pg_step}), 2) Metropolis-within-Gibbs (Section \ref{sec:mwg_step}), and 3) sampling importance resampling  (Section \ref{sec:sir_step}). These strategies are proposed in \citet{rezende2014stochastic,mattei2018leveraging,mattei2019miwae}, respectively, for missing  data imputation with deep generative models, and we describe a simple extension based on $\beta$-VAEs and the power likelihood.  Prior to imputation, we first train the $\beta$-VAE, using zero imputation for the missing values, to obtain estimates of generative model parameters $\btheta$ and variational parameters $\bphi$, and thus also an approximation of the true posterior of the latent variables.  

\subsubsection{Pseudo-Gibbs}\label{sec:pg_step}

Pseudo-Gibbs sampling was the first strategy developed to generate approximate samples from the predictive distribution in deep generative models   \citep{rezende2014stochastic}. In particular, approximate samples from the joint $p_{\btheta,\beta}(\bX_\mis, \bZ \mid \bX_\obs)$ are obtained by iteratively sampling from the encoder and decoder. More specifically, for
$s=1,\dots,S$ iterations and every data point $n \in \lbrace 1, \ldots, N \rbrace$ with missing features, the pseudo-Gibbs algorithm replaces the expectation steps in the single imputation of Section \ref{sec:singimp_qiu} with sampling, as follows: First, Sample $\bz_n$ (sample of encoder) given $\bx^{(s-1)}_{\mis,n}$:
    \begin{align}
    \bz_{n}^{(s)} \sim q_{\bphi}(\bz_n \mid \bx^{(s-1)}_{\mis,n}, \bx_{\obs,n}) .
\label{eq:encoder}        
    \end{align} 
Next, we sample  $\bx^{(s)}_{\mis,n}$ (sample of decoder) given $\bz^{(s)}_n$ based on the power likelihood: 
     $$ \bx^{(s)}_{\mis,n} \sim  p_{\btheta,\beta}(\bx_{\mis,n} \mid \bx_{\obs,n},  \bz^{(s)}_{n}) \propto p_{\btheta}(\bx_{\mis,n} \mid \bx_{\obs,n}, \bz^{(s)}_{n})^{1/\beta}. $$
Ideally, in the first step, we would aim to sample from the intractable true posterior of the latent variables. However, if the 
variational posterior provides a good approximation, 
the pseudo-Gibbs scheme will produce samples from a distribution close to our target.

\subsubsection{Metropolis-within-Gibbs}\label{sec:mwg_step}

The pseudo-Gibbs algorithm was improved and extended by \citet{mattei2018leveraging}, who derived a Metropolis-within-Gibbs (MWG) sampler that is asymptotically guaranteed to produce samples from the target predictive distribution. This is a simple modification of pseudo-Gibbs that corrects the first step by using the variational posterior as a proposal within a Metropolis-Hastings algorithm. Specifically, in the first step, the sampled value from the encoder in \eqref{eq:encoder} represents the proposed value for the latent variables, denoted by  $\bz_n^*$, which is then accepted according to the acceptance probability:
$$ a(\bz^{(s-1)}_n \rightarrow \bz^{*}_n) = \min\left(1, \frac{p_{\btheta}(\bx^{(s-1)}_{\mis,n}, \bx_{\obs,n}| \bz^*_n)^{1/\beta} p(\bz^*_n)}{p_{\btheta}(\bx^{(s-1)}_{\mis,n}, \bx_{\obs,n} |\bz^{(s-1)}_{n})^{1/\beta}p(\bz^{(s-1)})} \frac{q_{\bphi}(\bz^{(s-1)}_{n}| \bx^{(s-1)}_{\mis,n}, \bx_\obs )}{q_{\bphi}(\bz^{*}_n| \bx^{(s-1)}_{\mis,n}, \bx_{\obs,n} )}\right).$$ 
Thus, we set: 
 $$\bz^{(s)}_n = \left\lbrace\begin{array}{ll}
       \bz^{*}_n  & \text{ with prob. } a(\bz^{(s-1)}_n \rightarrow \bz^{*}_n)\\
       \bz^{(s-1)}_n  & \text{ with prob. } 1- a(\bz^{(s-1)}_n \rightarrow \bz^{*}_n)
    \end{array} \right. .$$ 
If the variational posterior is a perfect approximation of the true posterior, the acceptance probability will be one, and the algorithm reduces to pseudo-Gibbs.  In general, MWG acknowledges and corrects for the approximation of the posterior; however, if the variational posterior is far from the true posterior, MWG will suffer from low acceptance rates and slow convergence.  

\subsubsection{Sampling importance resampling}\label{sec:sir_step}

An alternative to Gibbs is sampling importance resampling (SIR), proposed by \citet{mattei2019miwae}. First, we perform importance sampling using the variational posterior as the importance distribution. In this case, for every data point $n \in \lbrace 1, \ldots, N \rbrace$ with missing features, we take $s=1,\ldots, S$ samples of the latent variables 
from our importance distribution: 
\begin{align*}
    \bz_n^{(s)} &\sim q_{\bphi}(\bz_n \mid \bx_{\mis,n}^{(0)},  \bx_{\obs,n}), 
\end{align*}
 where $\bx_{\mis,n}^{(0)}$ denotes an initial zero imputation for the missing data.
These importance samples $(\bz_n^{(s)})$, for $s=1,\ldots, S$, 
have weights $w^{(s)}_n$ proportional to:
\begin{align*}
    \omega^{(s)}_n = \frac{p_{\btheta}(\bx_{\obs,n}| \bz_n^{(s)})^{1/\beta}  p(\bz_n^{(s)})}{q_{\bphi}(\bz_n^{(s)}| \bx^{(0)}_{\mis,n}, \bx_{\obs,n} )}, 
\end{align*}
where $w^{(s)}_n = \omega^{(s)}_n / \sum_{s=1}^S \omega^{(s)}_n$ (for further details, see Appendix \ref{app:SIR}). 
Then, for multiple imputation, we obtain $M$ imputations by first sampling $(\bz_{n}^{m}) $, for $m=1, \ldots M$, with replacement from the importance samples $(\bz_{n}^{(s)}) $ 
with probability $w^{(s)}_n$. Next, for each  $\bz_{n}^{m} $, we impute the missing data by sampling from
$$\bx^{m}_{\mis,n} \sim p_{\btheta,\beta}(\bx_{\mis,n} \mid \bx_{\obs,n}, \bz^{m}_n).$$ In contrast to Gibbs sampling, an advantage of SIR is  parallelizability. However,  the discrepancy between the variational posterior and true posterior determines the efficiency of the algorithm, and a large discrepancy may result in degeneracy of the weights and  require a large number of importance samples (which is required to be exponential in KL divergence between the importance distribution and the target \citep{chatterjee2018sample}).   

\subsection{Cross-validation training regime}\label{sec:training_regime}

When the generative model and $\btheta$ match the true data generating distribution exactly, learning is achieved optimally with $\beta=1$. However, in practice, we have a mismatch and assigning a good value of $\beta$ becomes critical to achieve robustness and accurate uncertainty quantification. Indeed, if $\beta$ is set too low, the posterior uncertainty can be underestimated, while if $\beta$ is set too high, the posterior uncertainty is overestimated.  Some directions for assigning a value of $\beta$ from an information theoretic perspective are provided in \cite{holmes2017assigning}. Instead, we employ cross-validation to tune $\beta$ for accurate multiple imputation and coverage of the missing data. 

Specifically, the cross-validation approach to tuning $\beta$ and the number of epochs consists of creating $k$ copies of the data and adding a small proportion of additional missingness in each copy. We then carry out a grid search over the number of epochs and values of $\beta$, training $k$ models for each value of $\beta$. The final selection is the combination that has acceptable coverage while minimizing the mean absolute error (MAE) over the introduced missing values (averaged across the $k$ models). Once the optimal hyper-parameters for $\beta$ and epochs are selected, the model is retrained using all of the data. We observed that following this approach results in coverage and MAE on the test set being close to the values estimated through cross-validation. The introduction of additional missing values during cross-validation creates a slight bias towards selecting higher values of $\beta$ and fewer epochs but this bias is mitigated by increasing the number of copies, $k$, and thereby reducing the amount of additional missing values in each copy.

\subsection{Evaluating imputation performance}

To evaluate the imputation performance, we consider two quantities: 1) the mean absolute error (MAE) to assess  reconstruction accuracy and 2) the empirical coverage (EC) to quantify uncertainty.  
The MAE compares our imputed values to the ground truth that was originally masked in the complete dataset. Recall that $\widehat{\bX}_\mis = (\hat{\bx}_{\mis, 1},\dots,\hat{\bx}_{\mis,N})$ represents the imputed values, while $\bX_\mis$ represents the true (masked) values.
The MAE is defined as:
    \begin{equation}\label{eq:mae}
    \text{MAE} = \frac{1}{N}\sum_{n=1}^N | \widehat{\bx}_{\mis,n} - \bx_{\mis,n}|,
\end{equation}
where $| \widehat{\bx}_{\mis,n} - \bx_{\mis,n}|$ represents the average absolute difference across all missing features for the $n$th data point. For multiple imputation, the imputed values in \eqref{eq:mae} are averaged across the $M$ imputed datasets,  $\widehat{\bx}_{\mis,n} = \frac{1}{M}\sum_{m=1}^M \bx_{\mis,n}^m$. 
To evaluate uncertainty in multiple imputation, we 
first compute $100(1-\alpha)\%$ confidence intervals (CIs) for each missing value based on the $M$ imputed values. The empirical coverage is then computed as the fraction of 
times where the true value falls within the predicted interval. 

\section{Results: Genomic data imputation}\label{sec:results}

\subsection{Predictions are overconfident and missing data values are underestimated with single imputation and the standard VAE}\label{sec:res1}

\begin{figure}[!t]
\begin{center}
\subfloat[Accuracy at imputed values]{\includegraphics[width=0.3\textwidth]{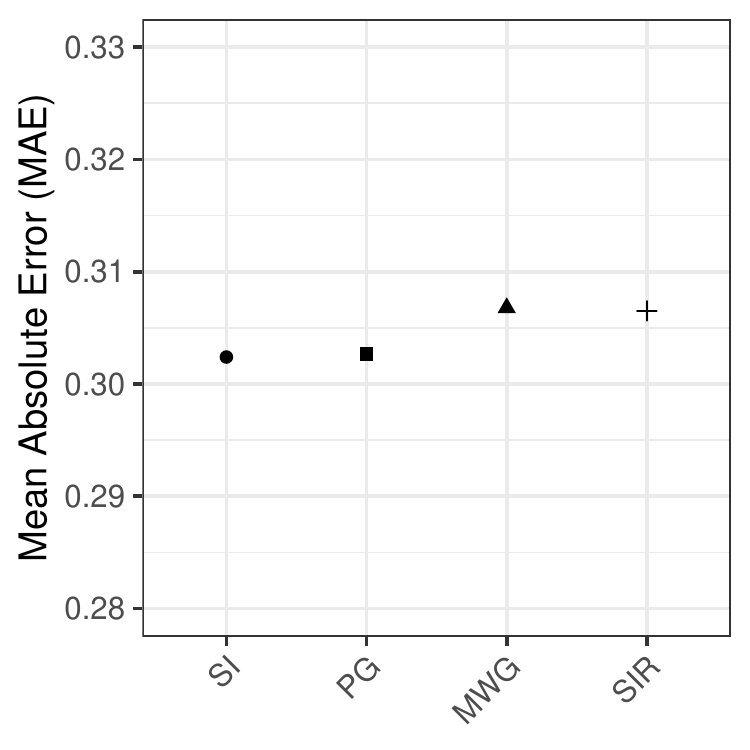}\label{fig:fig1a}}
\subfloat[Coverage at imputed values]{\includegraphics[width=0.3\textwidth]{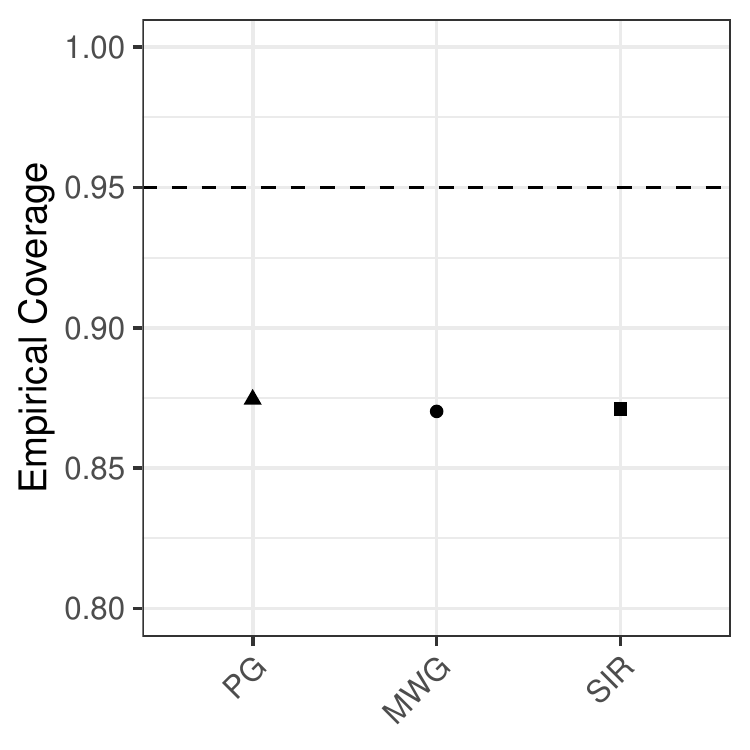}\label{fig:fig1b}}
\subfloat[Imputed values by SI]{\includegraphics[width=0.3\textwidth]{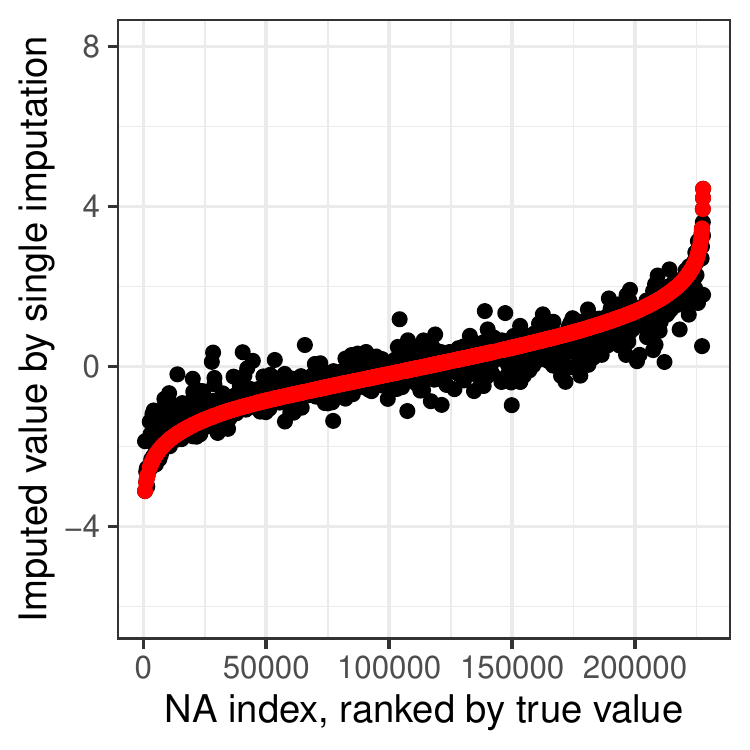}\label{fig:fig1c}}
\end{center}
\caption{\textbf{Standard VAE ($\mathbf{\beta}$=1) underestimates the uncertainty in multiple imputation.} Here we report \textbf{(a)} the accuracy of imputed missing data compared to the ground truth by MAE and \textbf{(b)} the fraction of true values that fall within the 95\% CIs 
for the three multiple imputation approaches pseudo-Gibbs (PG), Metropolis-within-Gibbs (MWG) and sampling importance resampling (SIR). Dotted line represents the desired coverage at 0.95. Finally, \textbf{(c)} depicts the imputed values for the missing data by single imputation (SI), ranked by their true values  (highlighted in red).}\label{fig:fig1}
\end{figure}

We first sought to investigate the impact of single imputation  with standard VAEs \citep{qiu2020genomic} compared to multiple imputation by our approach 
through Metropolis-within-Gibbs (MWG), pseudo-Gibbs (PG), and sampling importance resampling (SIR) by evaluating the accuracy of imputed values through MAE and assessing uncertainty through empirical coverage at 95\% CIs. We use the same publicly available RNA-sequencing dataset from the Cancer Genome Atlas (TCGA) used in \citet{qiu2020genomic} in order to benchmark our results against previous work done on genomic data imputation through VAEs. This dataset contains $D=17,175$ complete features for $N=667$ glioma patients, comprised of two cancer subtypes, glioblastoma (GBM) and low-grade glioma (LGG). We first simulate missingness in this dataset by masking values with 10\% missing completely at random (MCAR) in 20\% of samples, and subsequently train the VAE on all remaining observed data, $\bX_\obs$, with zeros imputed at missing value indices (see Section \ref{sec:methods}). In order to benchmark against their method, we use the same model and hyper-parameters that were found to be optimal in \citet{qiu2020genomic}, specifically, the standard VAE ($\beta$=1) with 250 training epochs and a learning rate of $10^{-5}$.
Once our model is trained, we generate $M=100$ plausible datasets for each multiple imputation approach and perform single imputation (as described in Section \ref{sec:singimp_qiu}). 

To evaluate imputation of the original masked values, we consider imputation accuracy by MAE and find that the multiple imputation approaches have similar accuracy to single imputation (SI), with pseudo-Gibbs performing slightly better than the other multiple imputation approaches (Figure \ref{fig:fig1a}). Next, we consider the empirical coverage of the masked values based on the 95\% CIs computed from the $M=100$ imputed datasets for all three multiple imputation approaches, and find that  the uncertainty is  underestimated (Figure \ref{fig:fig1b}). Additionally, the values imputed at masked data points with single imputation are underestimated at more extreme true values (Figure \ref{fig:fig1c}). This is common with neural networks, where model predictions can be poor but still reported with high confidence \citep{szegedy2013intriguing,nguyen2015deep}. Here we see that at more extreme true values, single imputation provides imputed values that are shifted towards the mean. This is likely due to overfitting of the trained model, resulting in overconfident predicted values for masked data points. As \citet{qiu2020genomic} only used reconstruction accuracy with single imputation to optimize hyperparameters, they were unable to assess uncertainty and overconfidence in the imputations.  
To overcome this, we explore regularization of the latent space through $\beta$-VAEs, optimizing the hyperparameters by considering both reconstruction accuracy and coverage. 

\subsection{Multiple imputation with $\beta$-VAE provides accurate uncertainty quantification while still retaining imputation accuracy}

\begin{figure}[!t]
\begin{center}
\subfloat[Single imputation]{\includegraphics[width=0.23\textwidth]{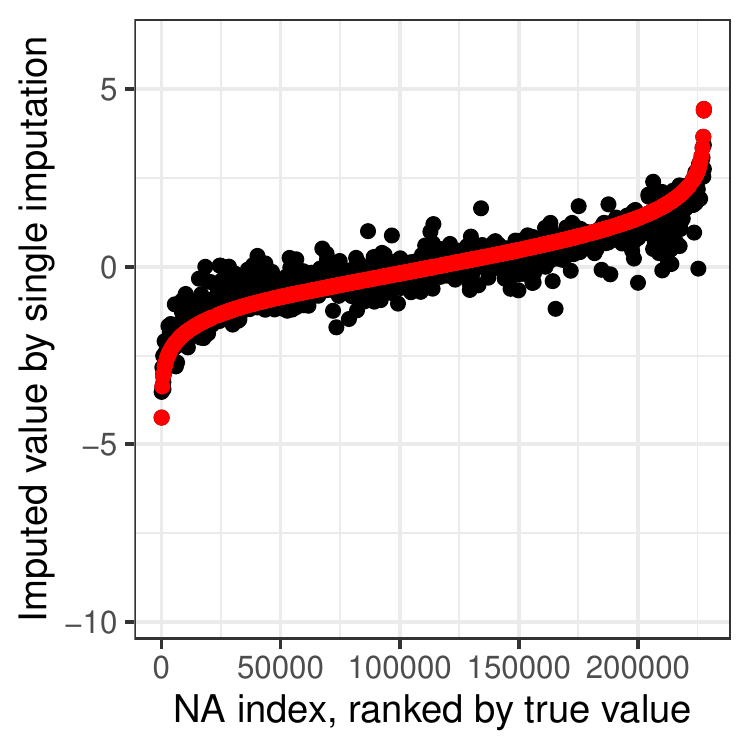}\label{fig:fig2a}}
\subfloat[PG]{\includegraphics[width=0.23\textwidth]{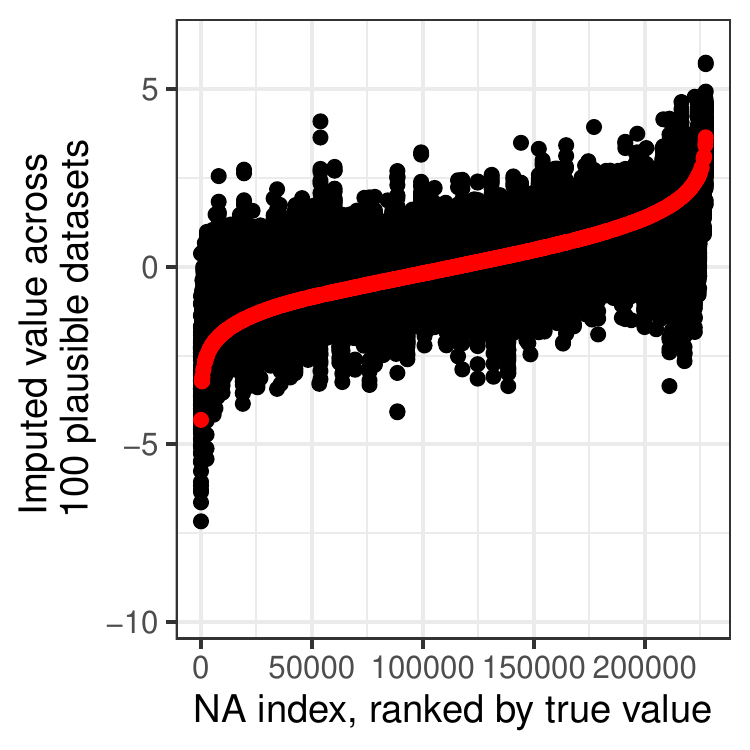}\label{fig:fig2b}}
\subfloat[MWG]{\includegraphics[width=0.23\textwidth]{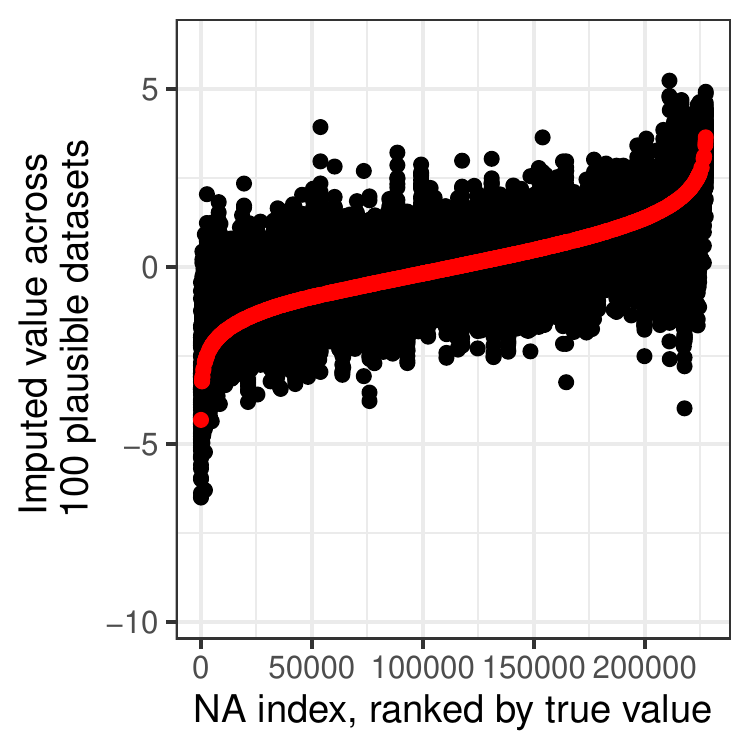}\label{fig:fig2c}}
\subfloat[SIR]{\includegraphics[width=0.23\textwidth]{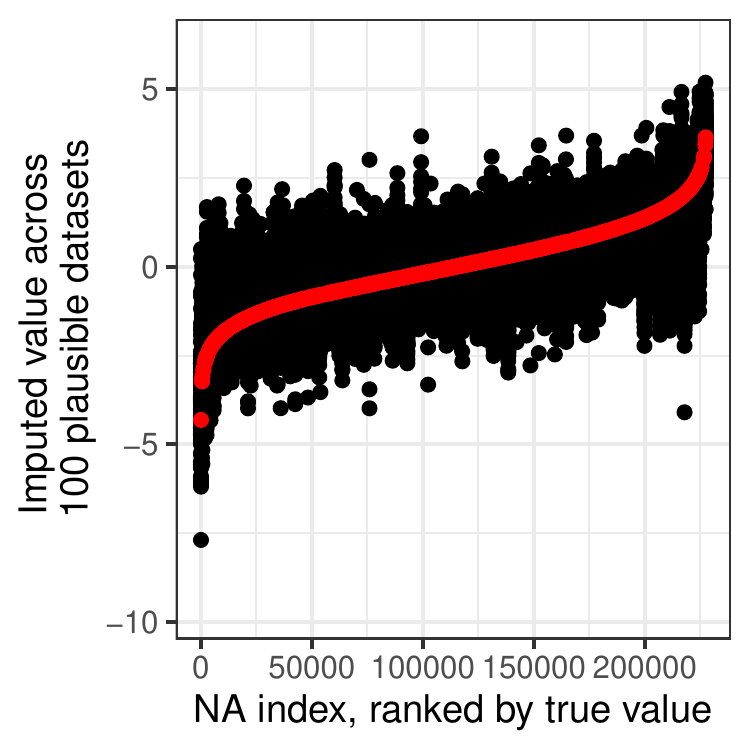}\label{fig:fig2d}} \\
\subfloat[Coverage at imputed values]{\includegraphics[width=0.46\textwidth]{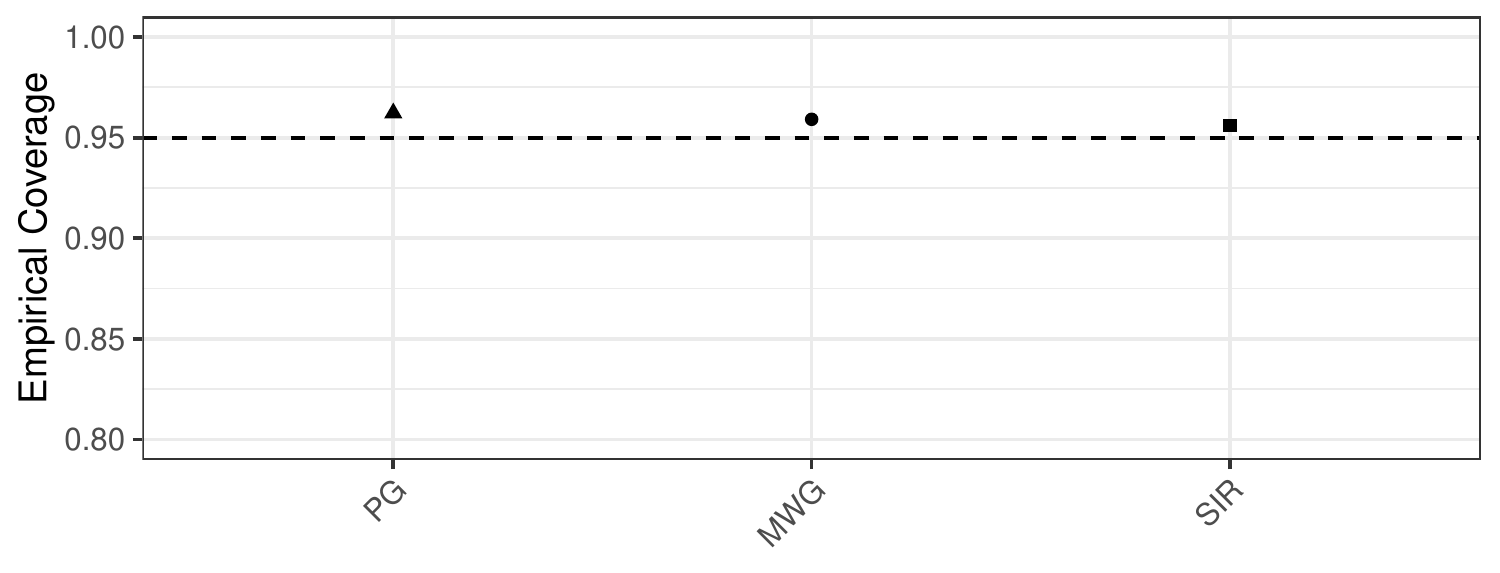}\label{fig:fig2e}}
\subfloat[Accuracy at imputed values]{\includegraphics[width=0.46\textwidth]{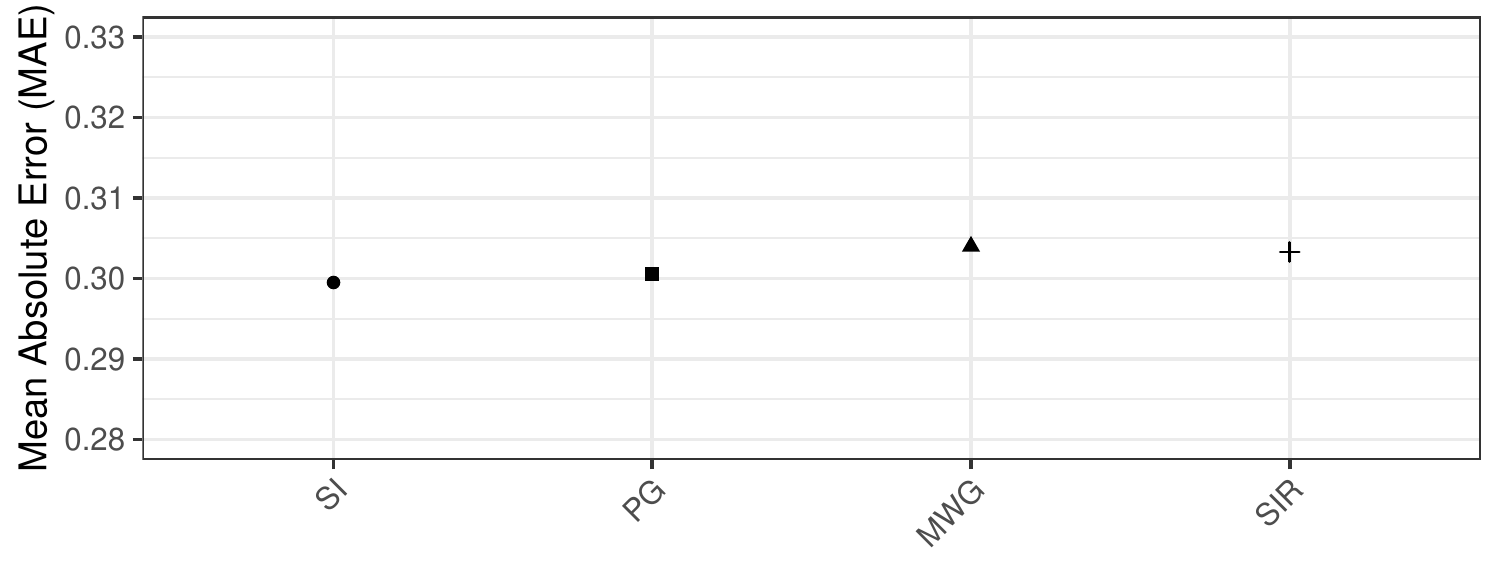}\label{fig:fig2f}}
\end{center}
\caption{\textbf{Multiple imputation with $\beta$-VAEs  provides proper coverage.} 
Here we report the (multiple) values imputed for the missing data, ranked by their true values (highlighted in red) for \textbf{(a)} single imputation (SI) and multiple imputation by \textbf{(b)} PG, \textbf{(c)}  MWG and \textbf{(d)} SIR. 
Imputation performance is summarized by \textbf{(e)} the empirical coverage at 95\% CIs (dotted line represents the desired coverage at 0.95) and \textbf{(f)} the accuracy at imputed values.}\label{fig:fig2}
\end{figure}

For improved robustness, we employ $\beta$-VAEs and the cross-validation scheme described in Section \ref{sec:training_regime} to tune $\beta$ and the number of training epochs, 
resulting in a value of $\beta = 2$ and 250 training epochs (Supplementary Figure \ref{sfig:sfig1}). We then train the $\beta$-VAE with these optimal parameters and impute values by single imputation and all three multiple imputation approaches PG, MWG and SIR. This results in good coverage at 95\%, with a much lower deviation from the desired coverage than the standard VAE with $\beta = 1$ (Figure \ref{fig:fig2a}-\ref{fig:fig2e}, Supplementary Figure \ref{sfig:sfig2}, Supplementary Figure \ref{sfig:sfig_percentiles}). Even with regularization of the latent space, single imputation still results in underestimation at extreme values (Figure \ref{fig:fig2a}). Our multiple imputation by $\beta$-VAEs yields good coverage across all missing data, even extreme values, while still retaining comparable accuracy to single imputation (Figure \ref{fig:fig2f}).

\subsection{Multiple imputation reduces false positive rate in downstream tasks. }

Lastly, we investigate the impact of all imputation approaches on downstream tasks, namely in identifying discriminating gene sets through logistic regression with the LASSO penalty. In particular,
we run LASSO regression on all imputed datasets to identify the genes which discriminate between the two cancer subtypes, GBM and LGG. This results in one gene set from our ground truth dataset with no missingness (GT), one from single imputation, and 100 discriminating gene sets for each multiple imputation approach, PG, MWG and SIR. 

We find that the union across all discriminating gene sets for each multiple imputation approach is much larger than the ground truth set, with the total number of possible non-zero coefficients ranging from 143 to 155, and only 31 discriminating genes in the true dataset (Table \ref{table:table1}). When comparing the estimated coefficients from the ground truth data to the (averaged) estimated coefficients based on the imputed data, this results in a slightly higher MAE for multiple imputation approaches (0.066, 0.064 and 0.069 for PG, MWG and SIR, respectively) compared to single imputation (0.053), which also has a set of 31 discriminating genes, although these are not identical to the ground truth set. However, when we inspect the coverage across the multiple imputations, we find that our coverage is close to the desired 95\% across PG, MWG and SIR (Table \ref{table:table1}, Supplementary Figure \ref{sfig:sfig3}). 

\begin{figure}[!t]
\begin{center}
\includegraphics[width=0.75\textwidth]{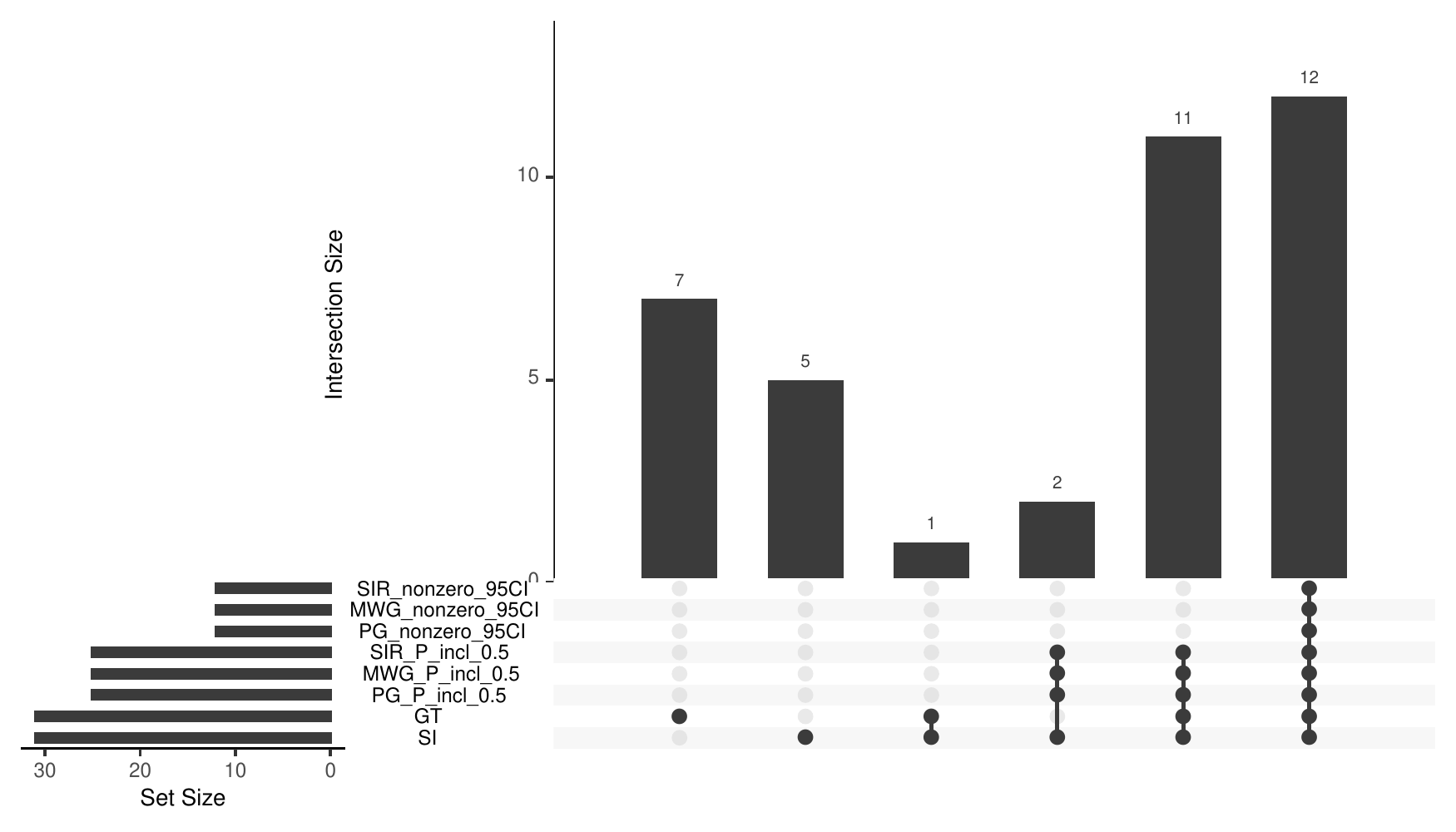}
\end{center}
\caption{\textbf{Upset plot showing overlapping discriminating gene sets by different imputation approaches.} We report the discriminating gene sets by single imputation (SI), ground truth (GT), and all three multiple imputation approaches PG, MWG and SIR with two different inclusion criteria, zero not contained in 95\% CI (nonzero\_95CI) and inclusion probability, $P_\text{incl} > 0.5$ (P\_incl\_0.5).}\label{fig:fig3}
\end{figure}

To identify discriminating gene sets across multiple imputations, we consider two approaches: selecting genes that 1) do not include a coefficient of zero in the 95\% CI computed from the  100 imputed datasets, and 2) have an inclusion probability, denoted  $P_{\text{incl}}$ and defined as the fraction of imputed datasets that the gene has a non-zero LASSO coefficient, greater than a specified threshold. 
The first approach results in the same set of 12 genes across all three multiple imputation approaches that are all in the true set of non-zero LASSO coefficients (Table \ref{table:table1}, Figure \ref{fig:fig3}), giving a false discovery rate (FDR) of 0\%. These 12 genes are also contained within the set for single imputation; however, single imputation results in 7 false positives
(Figure \ref{fig:fig3}), yielding an FDR of 22.6\% (7/31). 
In the second approach, if we threshold at $P_\text{incl} > 0.5$, this results in a final set of 25 discriminating genes for each multiple imputation approach (Table \ref{table:table1}, Supplementary Figure \ref{sfig:sfig4}). In this case,  our gene set contains 2 false positives, yielding an FDR of 8.0\% (Figure \ref{fig:fig3}). In summary, we find that multiple imputation with $\beta$-VAEs not only provides well-calibrated uncertainty but also results in much more acceptable FDRs in downstream tasks.  

\bgroup
\begin{table}[!t]
\centering
\caption{Performance of different imputation techniques, single imputation (SI) and multiple imputation by PG, MWG and SIR for imputation at missing value indices (first two rows) and downstream impact on LASSO regression (subsequent rows). The final row reports the false discovery rate, based genes with an inclusion probability $>0.5$ for multiple imputation.}
\label{table:table1}
\begin{tabular}{ |l||c|c|c|c|  }
 \hline
\textbf{Metric}& \textbf{SI}& \textbf{PG}& \textbf{MWG}& \textbf{SIR}  \\
 \hline
 \hline
MAE                 &  0.302   &  0.301   & 0.304   & 0.303\\
\hline
95\% CI coverage   & N/A    & 96.2\%    & 95.9\%    & 95.6\%\\
\hline
LASSO: MAE   & 0.053    & 0.066    & 0.064    & 0.069\\
\hline
LASSO: 95\% CI coverage    & N/A     &  97.4\% & 97.2\%    & 96.6\%      \\
\hline
LASSO: total number of non-zero coefficients    & 31     &  155 & 143    & 149      \\
\hline
LASSO: number of genes without zero in 95\% CI   &  N/A    &  12 & 12    & 12      \\
\hline
LASSO: number of genes with $P_\text{incl} > 0.5$    & N/A     &  25 & 25    & 25      \\
\hline
LASSO: False discovery rate   & 22.6\%     & 8\% & 8\%    & 8\%      \\
\hline
 \hline
\end{tabular}
\end{table}
\egroup

\section{Discussion}

We describe a deep learning framework for multiple imputation using $\beta$-VAEs.   We propose and compare three  multiple imputation methods and develop a new training regime, which uses all observed data to tune hyperparameters by assessing accuracy as well as empirical coverage. Our approach captures the complex, non-linear relationships present in high-dimensional genomic data, imputing values with high accuracy while retaining good coverage. Previous work \citep{qiu2020genomic} employed standard VAEs for genomic data imputation by single imputation, resulting in inaccurate and overconfident imputations at extreme missing values. 
Finally, we investigate the impact of these different imputation approaches on downstream tasks, namely discriminating gene sets identified by logistic regression with the LASSO penalty. We find that multiple imputation through $\beta$-VAEs identifies genes that discriminate between the two cancer subtypes with lower false discovery rates than previous methods. All three multiple imputation approaches perform similarly in terms of accuracy and coverage, however, SIR is preferable as it is much more computationally efficient. 

Future work will investigate  missing not at random settings \citep{ipsen2020not,collier2020vaes} and mixed data \citep{ma2020vaem}. In addition, extensions using ensembles of deep generative models may improve robustness and calibration. Such ensembles can be built from simple approaches, such as training with multiple initializations \citep{lakshminarayanan2017simple}, composing models across different epochs \citep{huang2017snapshot}, or Monte Carlo dropout \citep{gal2016dropout}, to more advanced approaches, such as Bayesian methods \citep{daxberger2019bayesian}.

\section*{Acknowledgements}
SW was supported by the Royal Society of Edinburgh (RSE) (grant number 69938). BRH and JW acknowledge the receipt of studentship awards from the Health Data Research UK-The Alan Turing Institute Wellcome PhD Programme in Health Data Science (Grant Ref: 218529/Z/19/Z).

\bibliography{iclr2023_conference}
\bibliographystyle{iclr2023_conference}

\appendix
\setcounter{table}{0}
\setcounter{figure}{0}
\renewcommand{\thetable}{\Alph{section}.\arabic{table}}
\renewcommand{\thefigure}{\Alph{section}.\arabic{figure}}
\renewcommand{\figurename}{Supplementary Figure}
\renewcommand{\tablename}{Supplementary Table}

\section{Appendix}

\subsection{Data and code availability}

All data used in this manuscript are publicly available. Gene expression data is version 2 of the adjusted pan-cancer gene expression data obtained from Synapse and can be found at \href{https://www.synapse.org/\#!Synapse:syn4976369.2}{https://www.synapse.org/\#!Synapse:syn4976369.2}. 

All code used to implement the analyses in this manuscript is hosted on GitHub at 
\href{https://github.com/roskamsh/BetaVAEMImputation}{https://github.com/roskamsh/BetaVAEMImputation} and will be made publicly available upon acceptance. 

\subsection{Software requirements}

The analyses carried out in this manuscript require the following software: python v3.10, TensorFlow v2.7.0; R: penalized v0.9, MASS v7.3, caret v6.0.

\subsection{$\beta$-VAEs and the power likelihood}\label{app:bvae} 

In the following, we show that the variational parameters $\bphi$ which maximize the $\beta$-VAE bound equivalently minimize the KL divergence between the variational posterior $q_{\bphi}(\bZ|\bX)$ and the true  posterior under the power likelihood. Specifically, the KL divergence between the variational posterior and the posterior under the power likelihood is given by:
\begin{align*}
 &D_{\text{KL}}(q_{\bphi}(\bZ|\bX),p_{\btheta,\beta}(\bZ|\bX)) = E_{\bZ \sim q_{\bphi}(\bZ|\bX)} \left[ \log \left( \frac{p_{\btheta,\beta}(\bX) q_{\bphi}(\bZ|\bX)}{p_{\btheta}(\bX \mid \bZ)^{1/\beta} p(\bZ)} \right) \right]   \\
 &\quad \quad = -\sum_{n=1}^N E_{\bz_n \sim q_{\bphi}(\bz_n|\bx_n)} \left[ \log \left( p_{\btheta}(\bx_n \mid \bz_n)^{1/\beta} \right) \right] + \sum_{n=1}^N E_{\bz_n \sim q_{\bphi}(\bz_n|\bx_n)} \left[ \log \left( \frac{ q_{\bphi}(\bz_n|\bx_n)}{p(\bz_n)} \right) \right] \\
 &\quad \quad \quad \quad + \log\left( p_{\btheta,\beta}(\bX)\right)\\
 &\quad \quad = \text{const.}  -\sum_{n=1}^N E_{\bz_n \sim q_{\bphi}(\bz_n|\bx_n)} \left[ \log \left( p_{\btheta}(\bx_n \mid \bz_n)^{1/\beta} \right) \right] + D_{\text{KL}}(q_{\bphi}(\bz_n|\bx_n),p(\bz_n)). 
\end{align*}
Thus, we can equivalently find $\bphi$, which maximize the ELBO:
\begin{align*}
    \text{ELBO} = \sum_{n=1}^N\E_{\bz_n \sim q_{\bphi}(\bz_n|\bx_n)}[\log{p_{\btheta}(\bx_n|\bz_n)}] - \beta \, D_{\text{KL}}(q_{\bphi}(\bz_n|\bx_n),p(\bz_n)).
\end{align*}

\paragraph{Example: factorized Gaussian.} Assume the generative model is a factorized Gaussian (as is used for the genomic data in Section \ref{sec:results}):
\begin{align*}
   p_{\btheta}(\bx_n|\bz_n) &=  \prod_{d=1}^D \Norm\left( x_{n,d} \mid \mu_{d}(\bz_n), \sigma^2_d(\bz_n) \right),
\end{align*}
where $(\mu_{d}(\bz_n), \sigma^2_d(\bz_n))$ for $d=1,\ldots, D$ represent the output of the final layer of the neural network with weights and biases contained in $\btheta$. In this case, the full conditional of the missing data under the power likelihood is
\begin{align*}
p_{\btheta,\beta}(\bx_{\mis,n} \mid \bx_{\obs,n},  \bz_n) &\propto 
   p_{\btheta}(\bx_{\mis,n} \mid \bx_{\obs,n},  \bz_n)^{1/\beta} 
   \\
   &=  \left(\prod_{d \in \mathcal{D}_{\mis,n}} \Norm\left( x_{n,d} \mid \mu_{d}(\bz_n), \sigma^2_d(\bz_n) \right) \right)^{1/\beta}\\
   &=  \left(\prod_{d \in \mathcal{D}_{\mis,n}} \frac{1}{\sqrt{2\pi \sigma^2_d(\bz_n)}} \exp\left( \frac{1}{2 \sigma^2_d(\bz_n) } (x_{n,d}- \mu_{d}(\bz_n))^2 \right) \right)^{1/\beta}\\
   &\propto  \prod_{d \in \mathcal{D}_{\mis,n}} \exp\left( \frac{1}{2 \beta \sigma^2_d(\bz_n) } (x_{n,d}- \mu_{d}(\bz_n))^2 \right) \\
   &\propto \prod_{d \in \mathcal{D}_{\mis,n}} \Norm\left( x_{n,d} \mid \mu_{d}(\bz_n), \beta \sigma^2_d(\bz_n) \right), 
\end{align*}
where $\mathcal{D}_{\mis,n} \subseteq \lbrace 1, \ldots, D \rbrace$ contains the indices of the missing features for the $n$th data point. Thus, in this case, sampling from the full conditional of the missing data under the power likelihood corresponds to sampling from the Gaussian with variance rescaled by a factor of $\beta$. Note that for $\beta>1$ this corresponds to increasing the spread and uncertainty of the missing data, which is critical to improve coverage of the deep generative model. 

\subsection{Sample importance resampling}\label{app:SIR}

We first note that our SIR scheme differs slightly from the scheme proposed by \cite{mattei2019miwae}, who  propose joint importance samples  $(\bz_n^{(s)}, \bx^{(s)}_{\mis,n})$ from
\begin{align*}
    \bz_n^{(s)} &\sim q_{\bphi}(\bz_n \mid \bx_{\mis,n}^{(0)},  \bx_{\obs,n}), \quad
    \bx^{(s)}_{\mis,n} \sim p_{\btheta}(\bx_{\mis,n} \mid \bx_{\obs,n}, \bz^{(s)}_n).
\end{align*}
Instead, we only consider importance sampling for  $(\bz_n^{(s)})$ and subsequently sample the missing data for each of resampled latent variables. Importantly, if the effective sample size is low, resulting in potential duplicates in the $M$ samples of latent variables, we obtain improved variability across multiple imputations of the missing data, compared to the approach of \cite{mattei2019miwae}.  

\paragraph{Example: factorized Gaussian.} Assume the generative model is a factorized Gaussian, then the importance weights are proportional to :
\begin{align*}
    \omega^{(s)}_n &= \frac{p_{\btheta}(\bx_{\obs,n}| \bz_n^{(s)})^{1/\beta}  p(\bz_n^{(s)})}{q_{\bphi}(\bz_n^{(s)}| \bx^{(0)}_{\mis,n}, \bx_{\obs,n} )}\\
    &= \frac{\left(\prod_{d \in \mathcal{D}_{\obs,n}} \Norm\left( x_{n,d} \mid \mu_{d}(\bz_n), \sigma^2_d(\bz_n) \right) \right)^{1/\beta}\Norm(\bz_n^{(s)} \mid \mathbf{0}, \mathbf{I})}{q_{\bphi}(\bz_n^{(s)}| \bx^{(0)}_{\mis,n}, \bx_{\obs,n} )},
\end{align*}
where $\mathcal{D}_{\obs,n} \subseteq \lbrace 1, \ldots, D \rbrace$ contains the indices of the observed features for the $n$th data point.

\subsection{Supplemental Figures}

\begin{figure}[!h]
\begin{center}
{\includegraphics[width=1.0\textwidth]{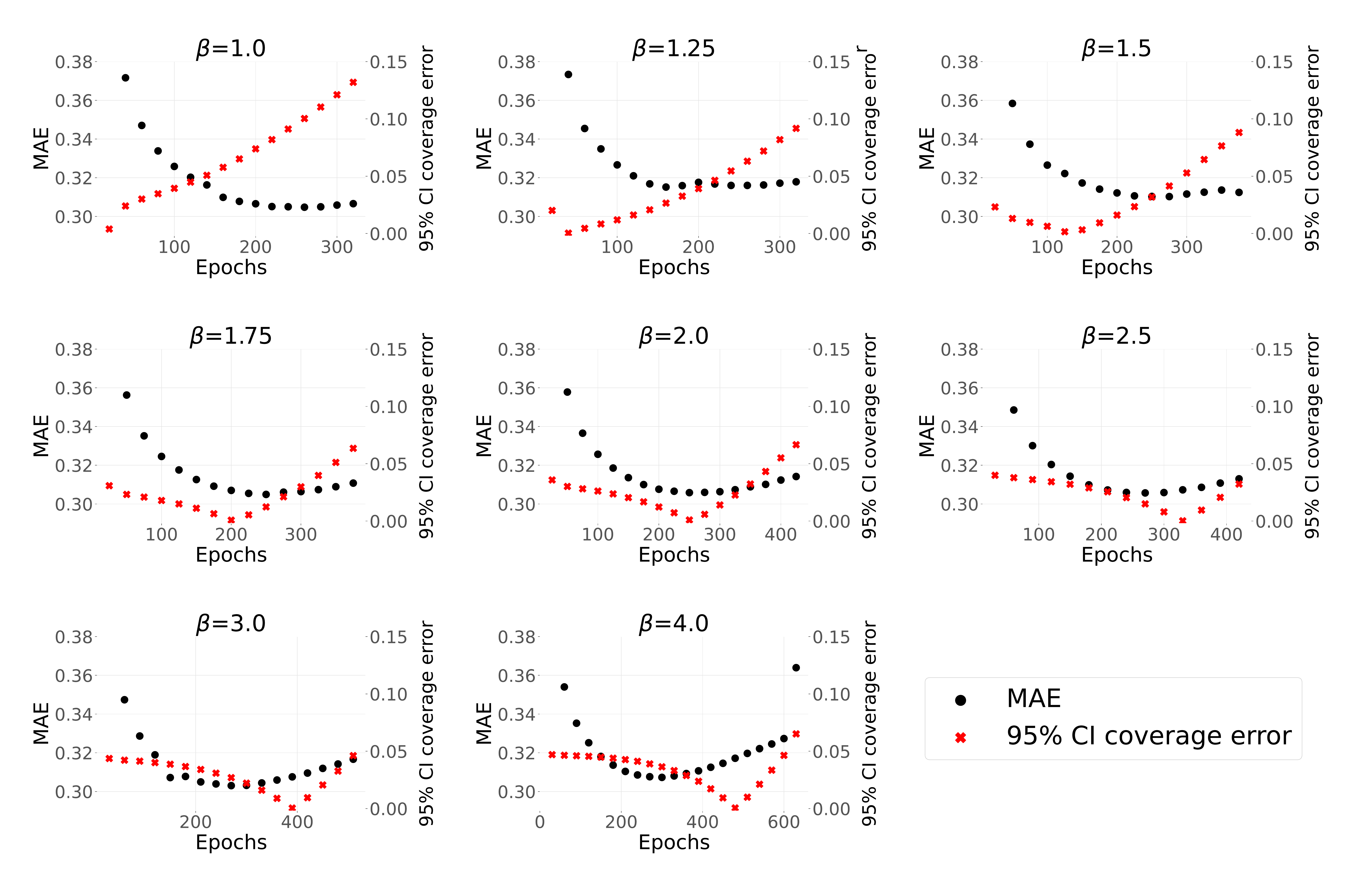}\label{sfig:sfig1a}} 
\end{center}
\caption{\textbf{Results from 5-fold cross-validation to determine optimal model and hyper-parameters.} MAE (black) and EC (red) of 95\% CI (computed based on quantiles) at different training epochs. We aim to minimize both of these metrics for optimal training parameters.}\label{sfig:sfig1}
\end{figure}

\begin{figure}[!h]
\begin{center}
\subfloat[SI]{\includegraphics[width=0.3\textwidth]{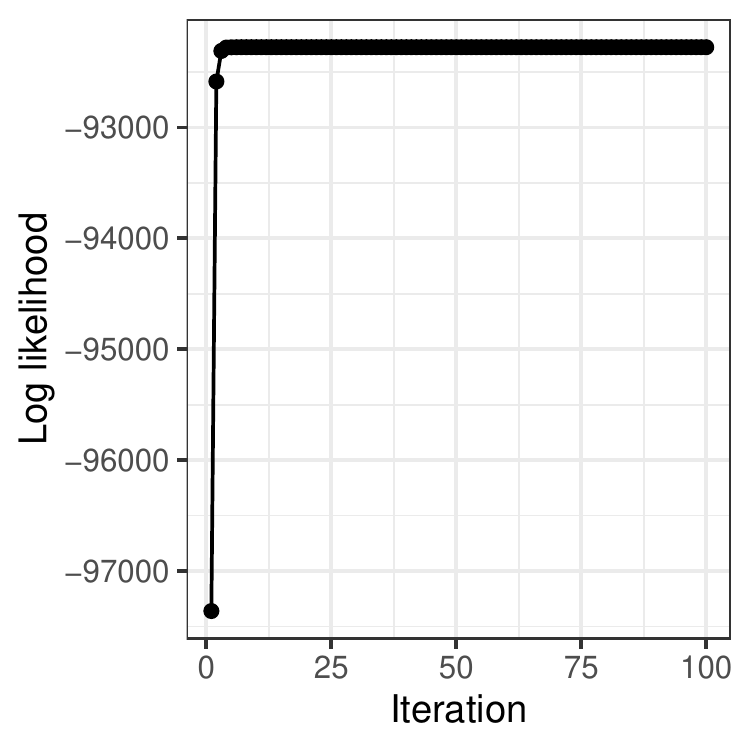}\label{sfig:sfig2a}}
\subfloat[PG]{\includegraphics[width=0.3\textwidth]{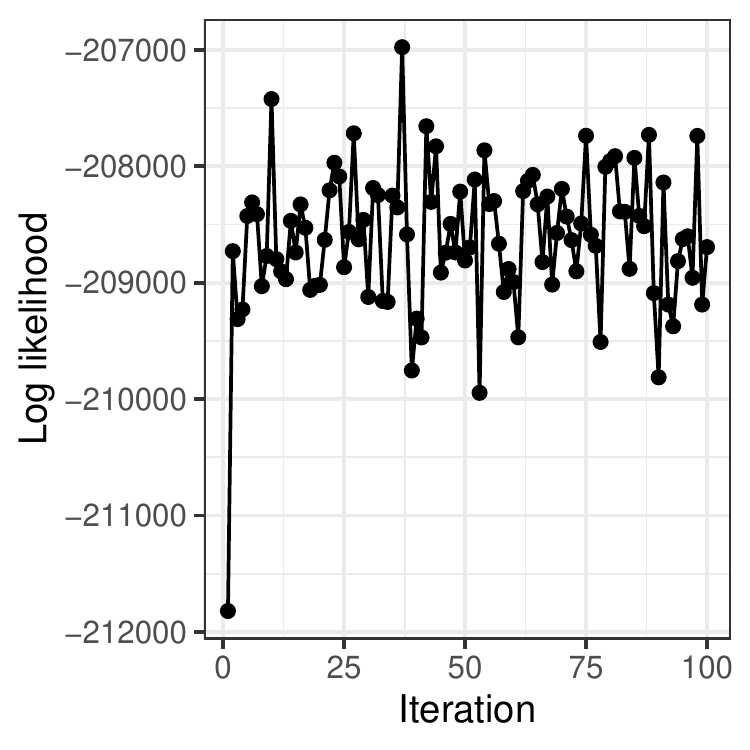}\label{sfig:sfig2b}}
\subfloat[MWG]{\includegraphics[width=0.3\textwidth]{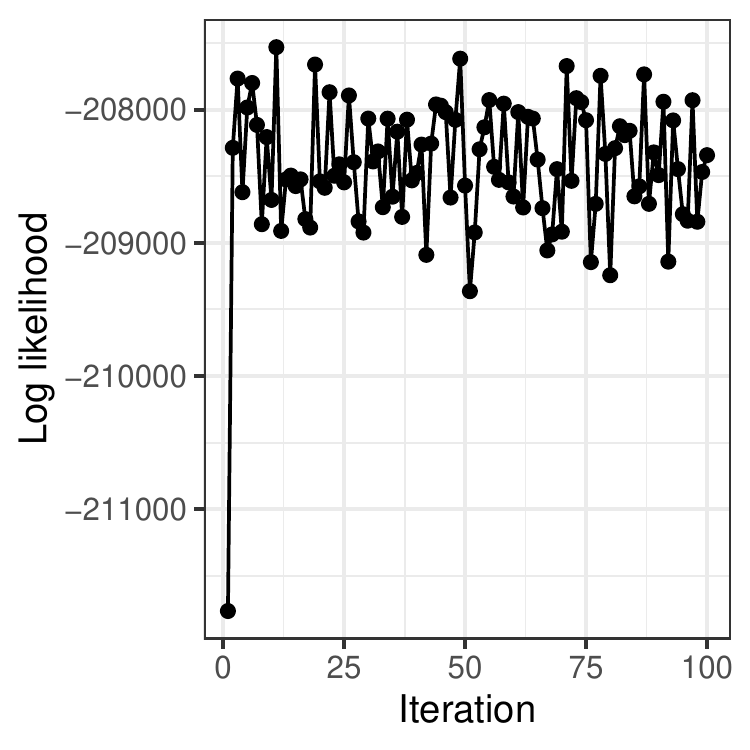}\label{sfig:sfig2b}}
\end{center}
\caption{\textbf{Trace plots monitoring convergence of Markov Chain Monte Carlo schemes for $\beta=2$ case.} Here we report the log likelihood of the data under the generative model at each iteration for \textbf{(a)} single imputation (SI), \textbf{(b)} pseudo-Gibbs (PG) and \textbf{(c)} Metropolis-within-Gibbs (MWG). For visualization purposes, we show iterations 1 to 100, but ran to 1000 iterations in implementation.}\label{sfig:trace_plots}
\end{figure}

\begin{figure}[!h]
\begin{center}
\subfloat[$\beta=1$]{\includegraphics[width=0.4\textwidth]{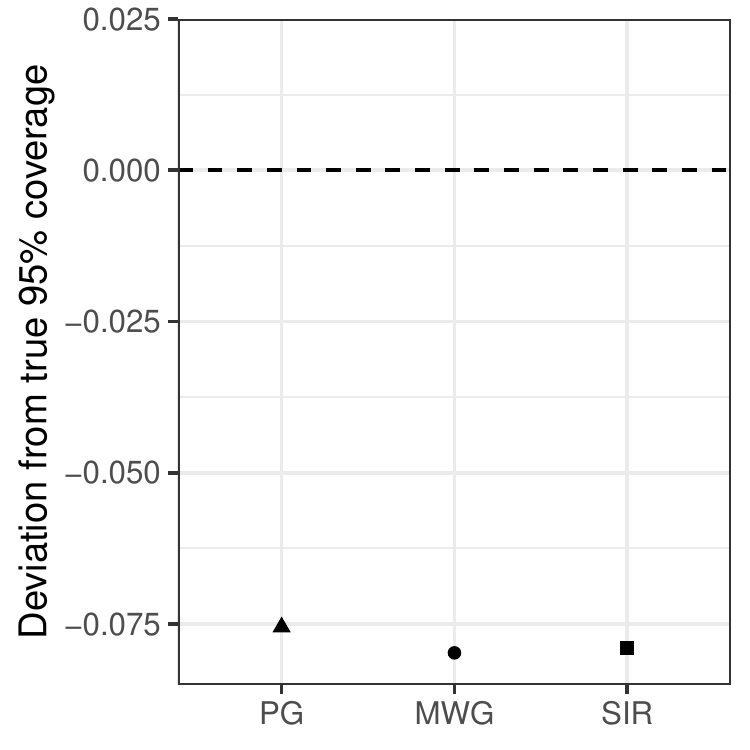}\label{sfig:sfig2a}}
\subfloat[$\beta=2$]{\includegraphics[width=0.4\textwidth]{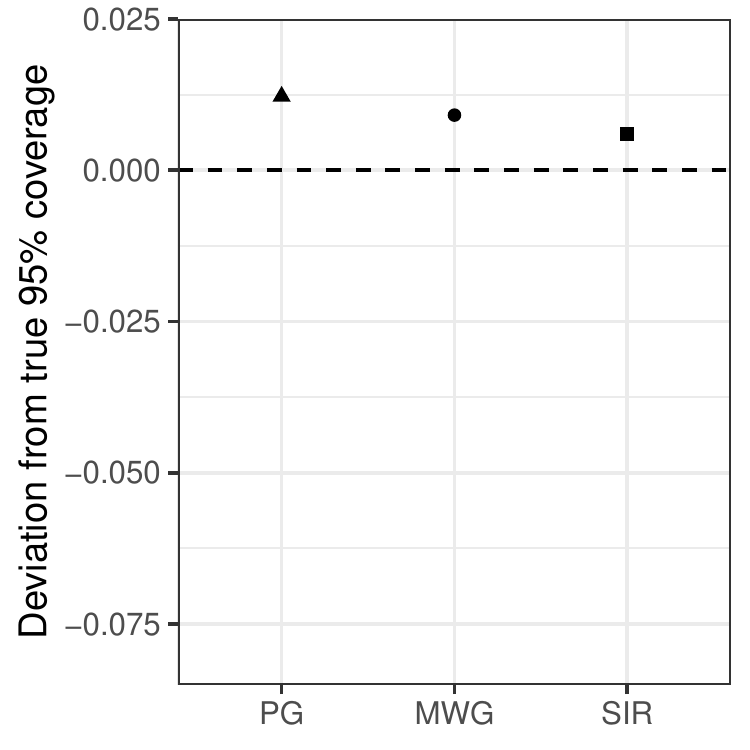}\label{sfig:sfig2b}}
\end{center}
\caption{\textbf{Deviation from true coverage for all three multiple imputation approaches.} Here we report the deviation from the desired coverage of 95\% (showed here by the dotted line) for all three multiple imputation approaches pseudo-Gibbs (PG), Metropolis-within-Gibbs (MWG) and sampling importance resampling (SIR) for \textbf{(a)} $\beta=1$, and \textbf{(b)} $\beta=2$.}\label{sfig:sfig2}
\end{figure}

\begin{figure}[!h]
\begin{center}
\subfloat[$\beta=1$]{\includegraphics[width=0.5\textwidth]{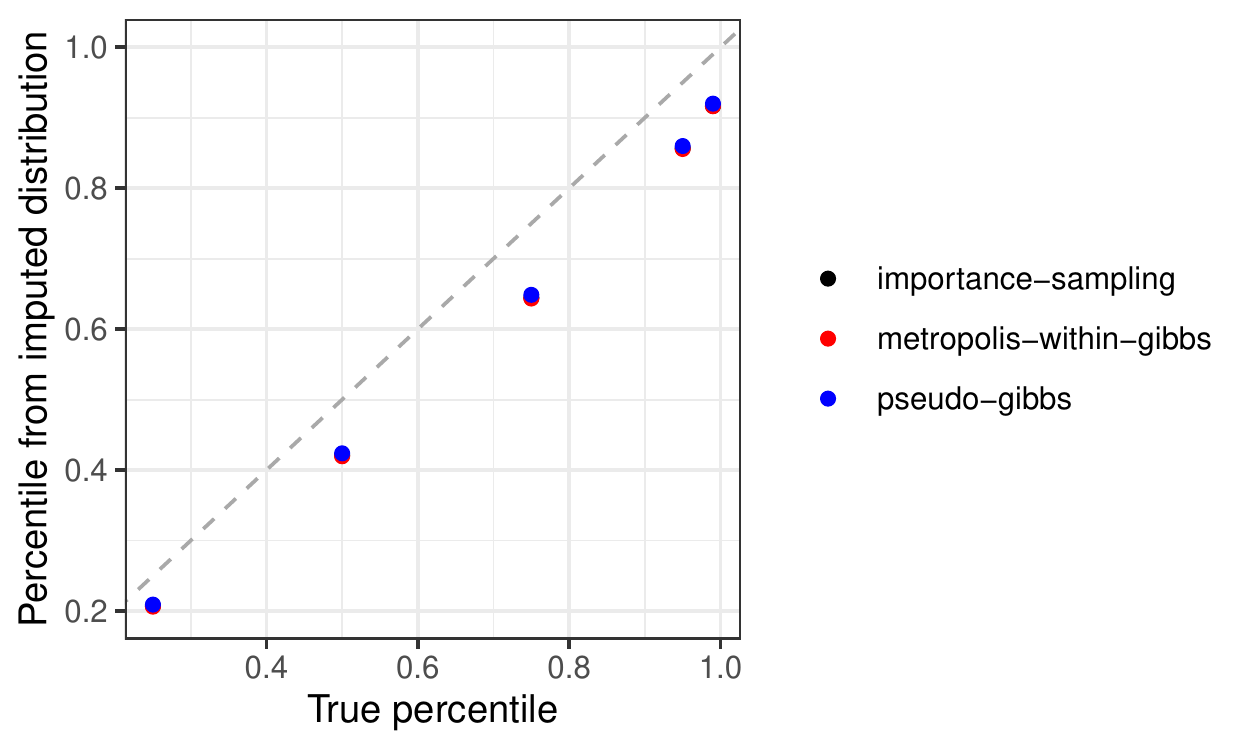}\label{sfig:sfig2a}}
\subfloat[$\beta=2$]{\includegraphics[width=0.5\textwidth]{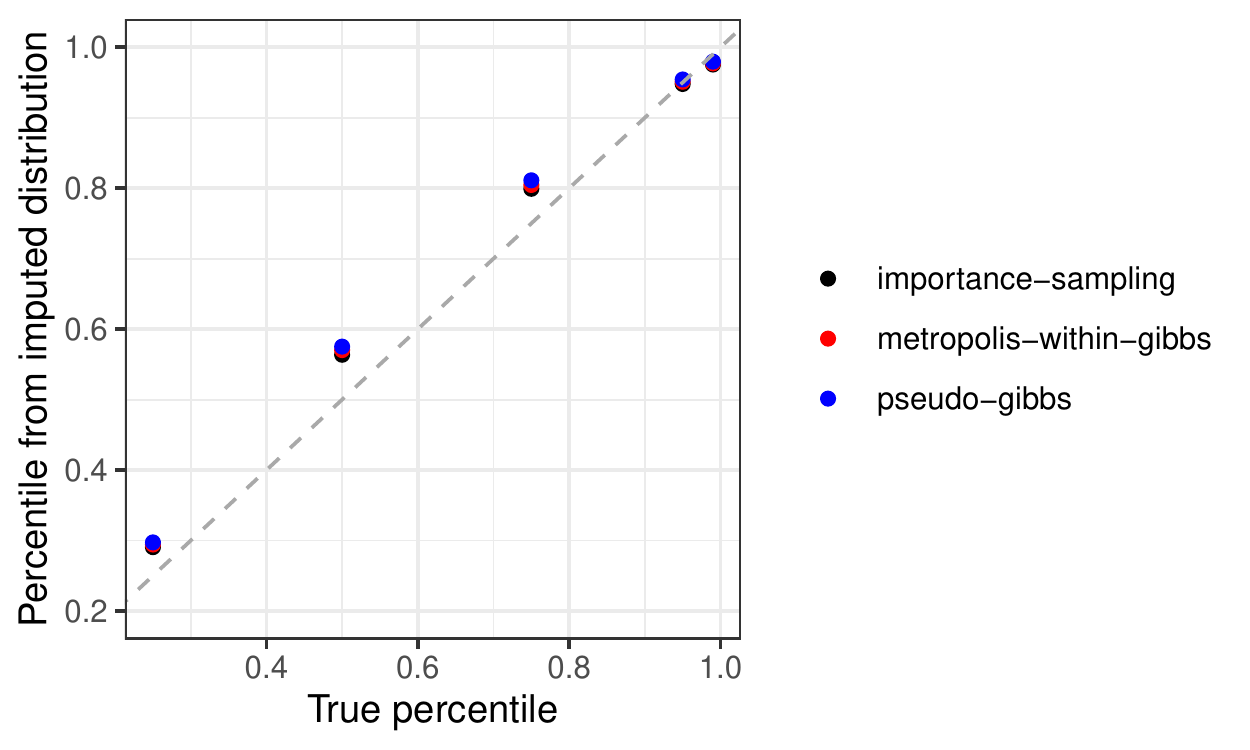}\label{sfig:sfig2b}}
\end{center}
\caption{\textbf{Coverage by percentiles of all three multiple imputation approaches.} Here we report the coverage evaluated by percentiles across all imputed datasets compared to the true percentiles of 0.25, 0.5, 0.75, 0.95 and 0.99 for all three multiple imputation approaches pseudo-Gibbs (PG), Metropolis-within-Gibbs (MWG) and sampling importance resampling (SIR) for \textbf{(a)} $\beta=1$, and \textbf{(b)} $\beta=2$.}\label{sfig:sfig_percentiles}
\end{figure}

\begin{figure}[!h]
\begin{center}
\subfloat[PG]{\includegraphics[width=0.9\textwidth]{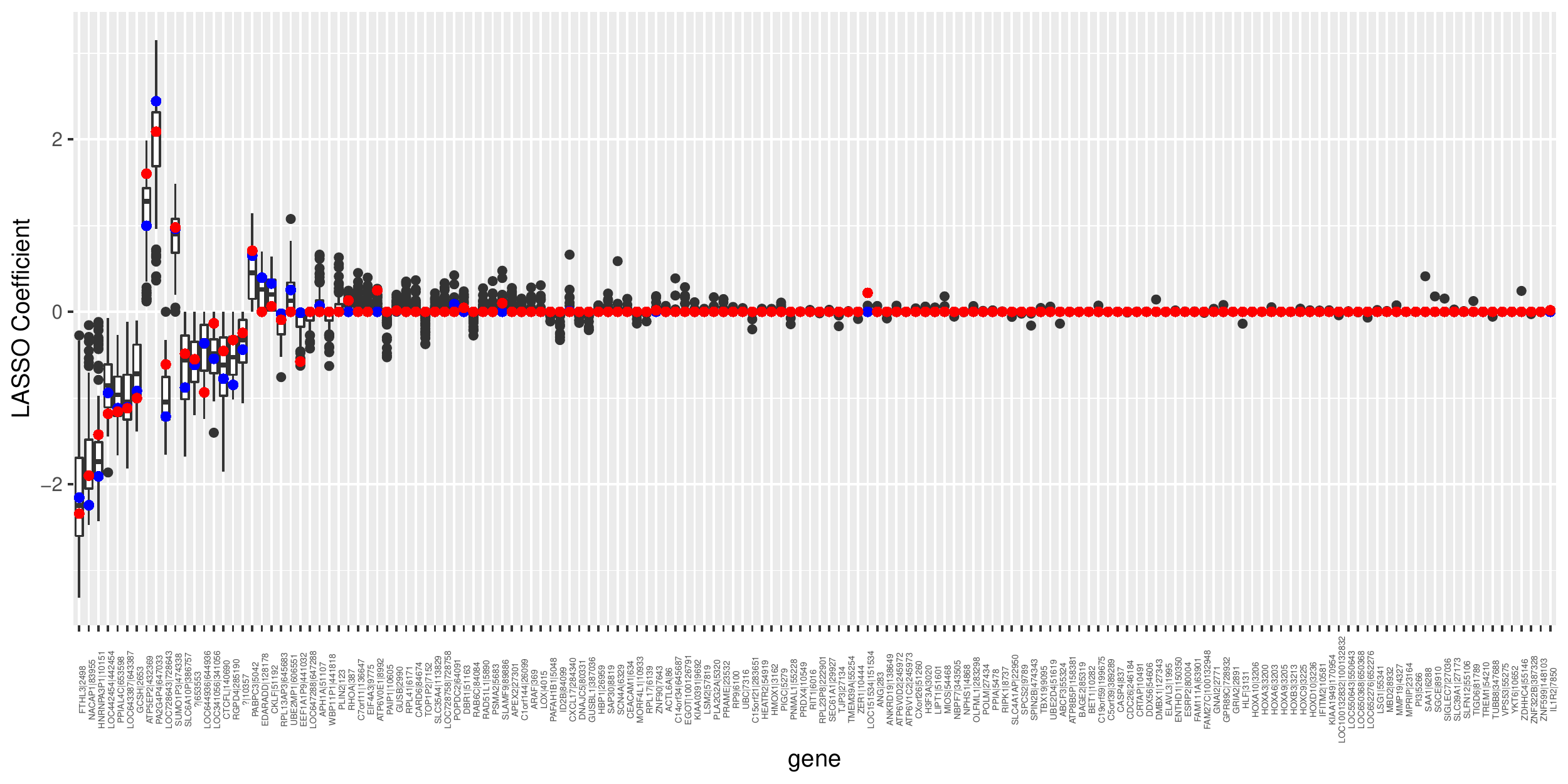}\label{sfig:sfig3a}} \\
\subfloat[MWG]{\includegraphics[width=0.9\textwidth]{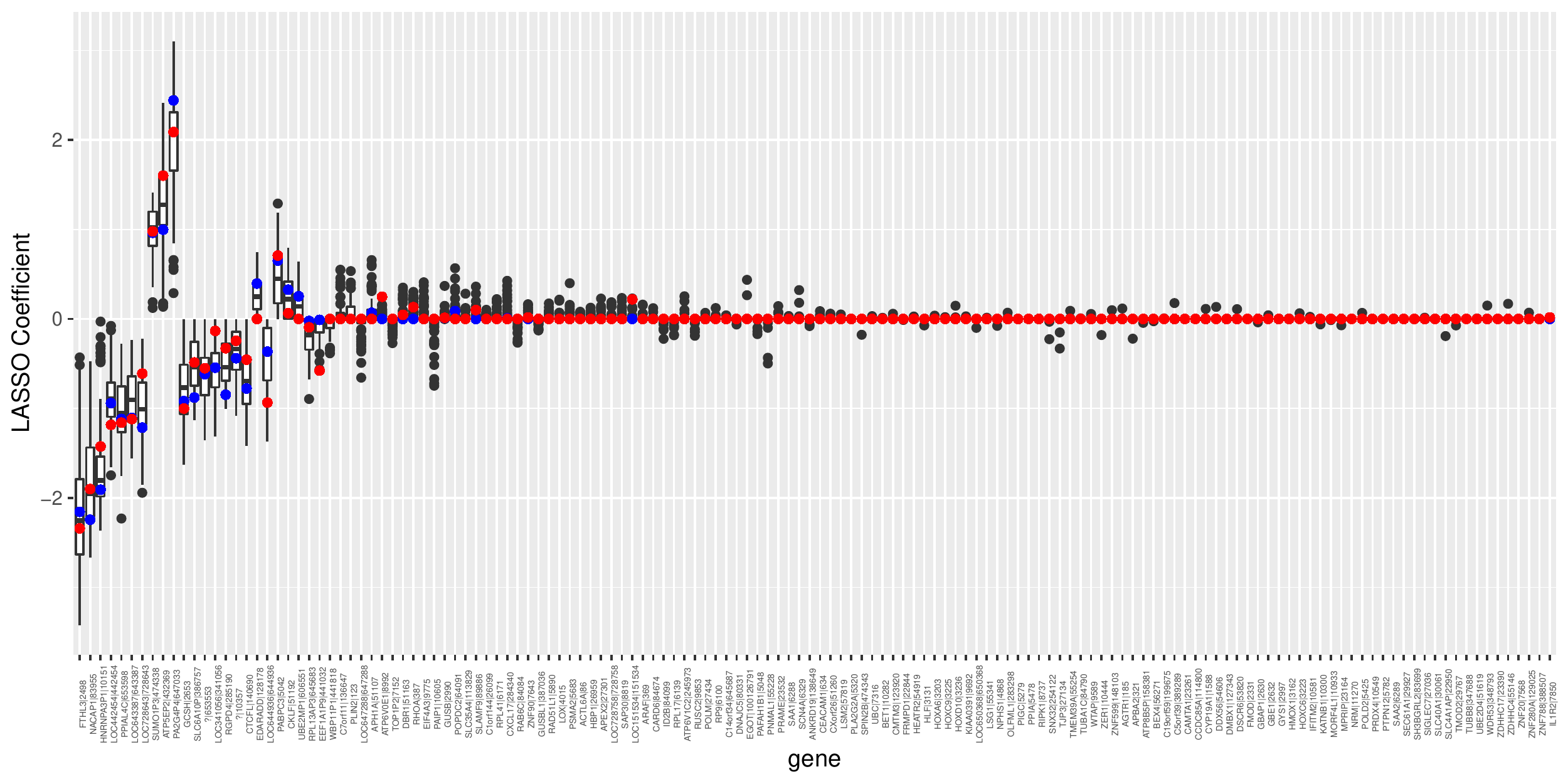}\label{sfig:sfig3b}} \\
\subfloat[SIR]{\includegraphics[width=0.9\textwidth]{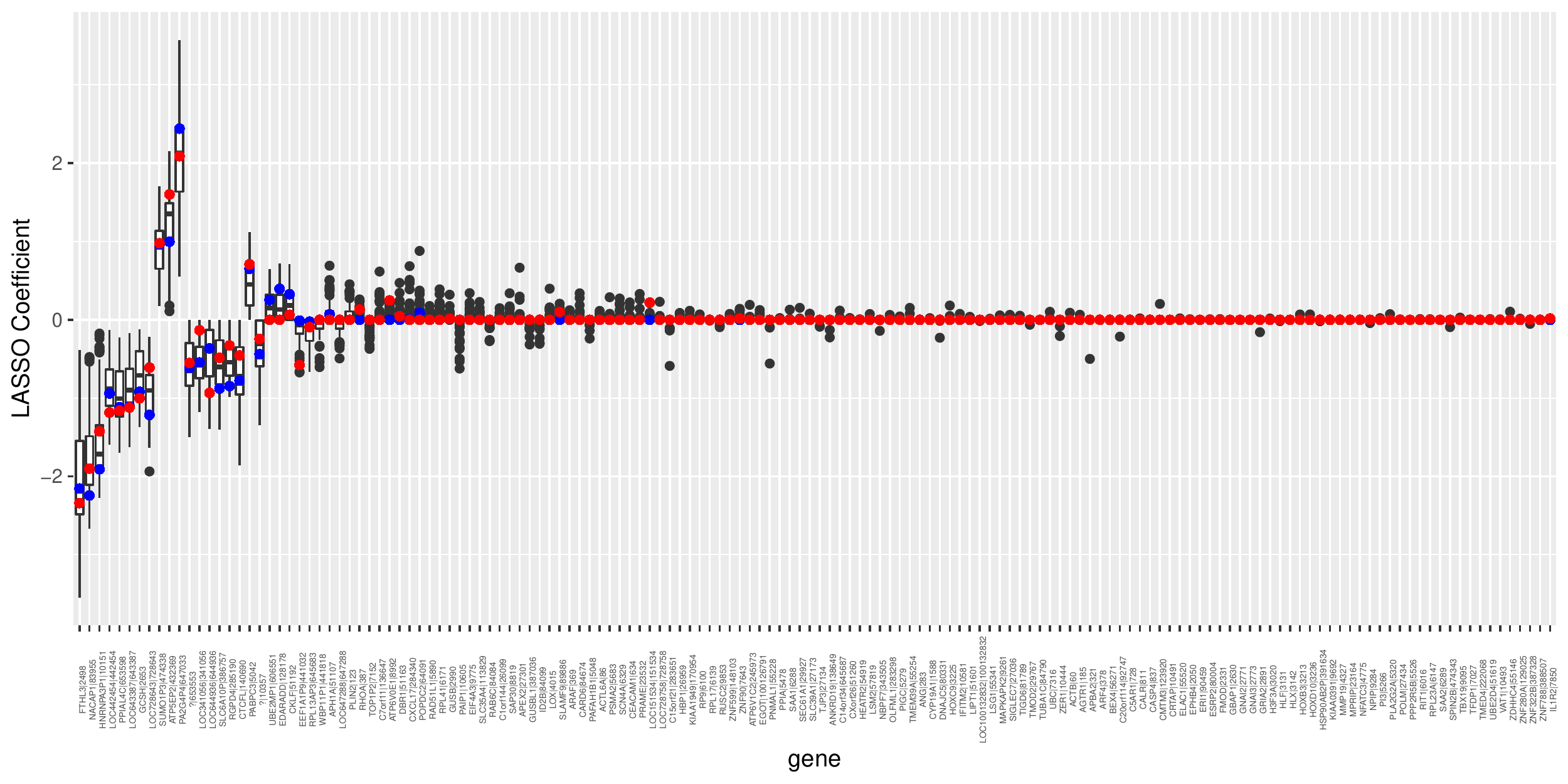}\label{sfig:sfig3c}} \\
\end{center}
\caption{\textbf{LASSO regression coefficients across multiple imputation approaches.} Here we report the LASSO regression coefficients across all 100 plausible imputed datasets for \textbf{(a)} pseudo-Gibbs (PG), \textbf{(b)} Metropolis-within-Gibbs (MWG) and \textbf{(c)} sampling importance resampling (SIR). LASSO regression coefficient value for the true dataset is highlighted in red, and for single imputation is highlighted in blue. For the purpose of visualization, the intercept was removed from this plot.}\label{sfig:sfig3}
\end{figure}

\begin{figure}[!h]
\begin{center}
\subfloat[PG]{\includegraphics[width=0.9\textwidth]{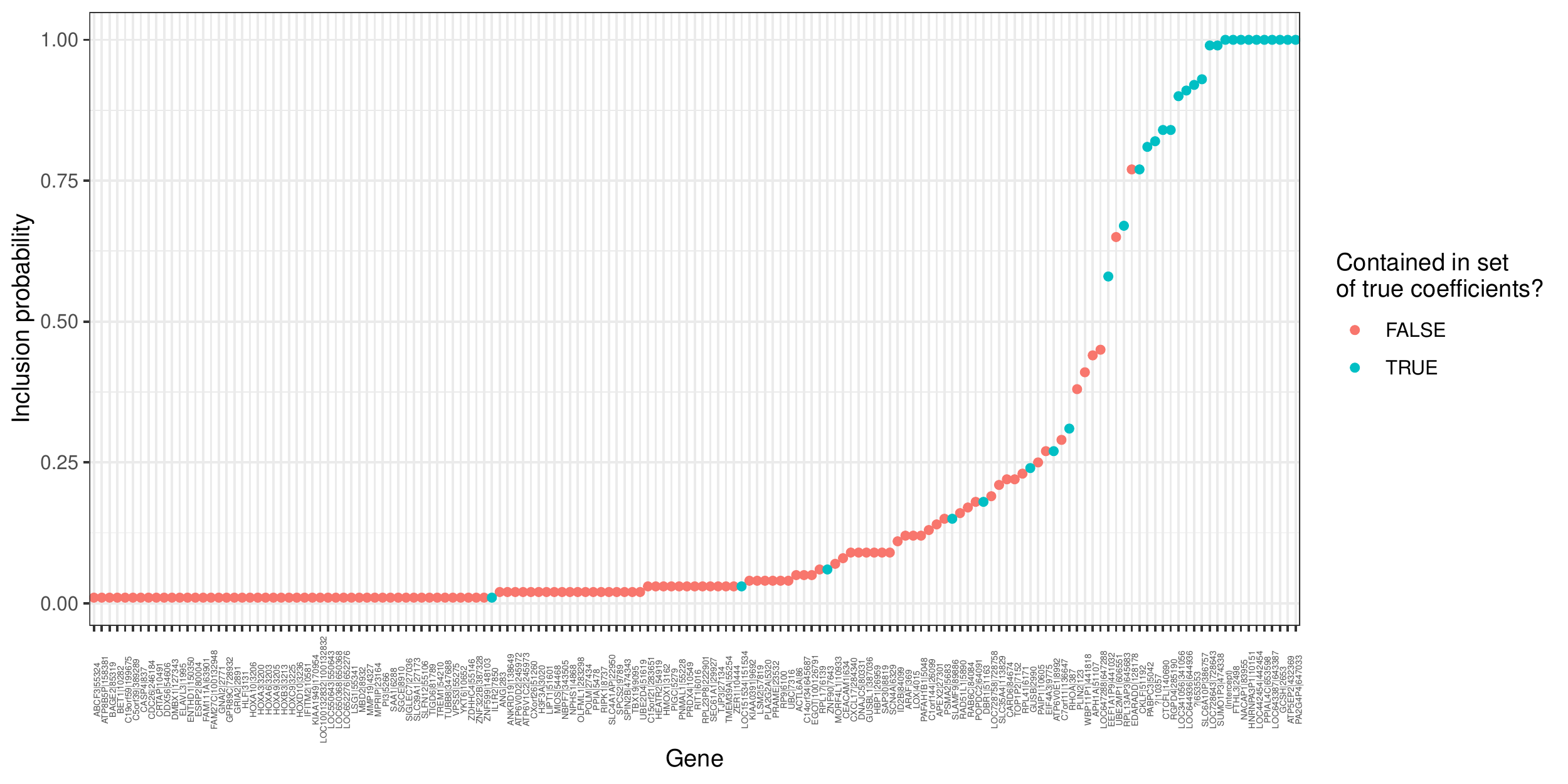}\label{sfig:sfig4a}} \\
\subfloat[MWG]{\includegraphics[width=0.9\textwidth]{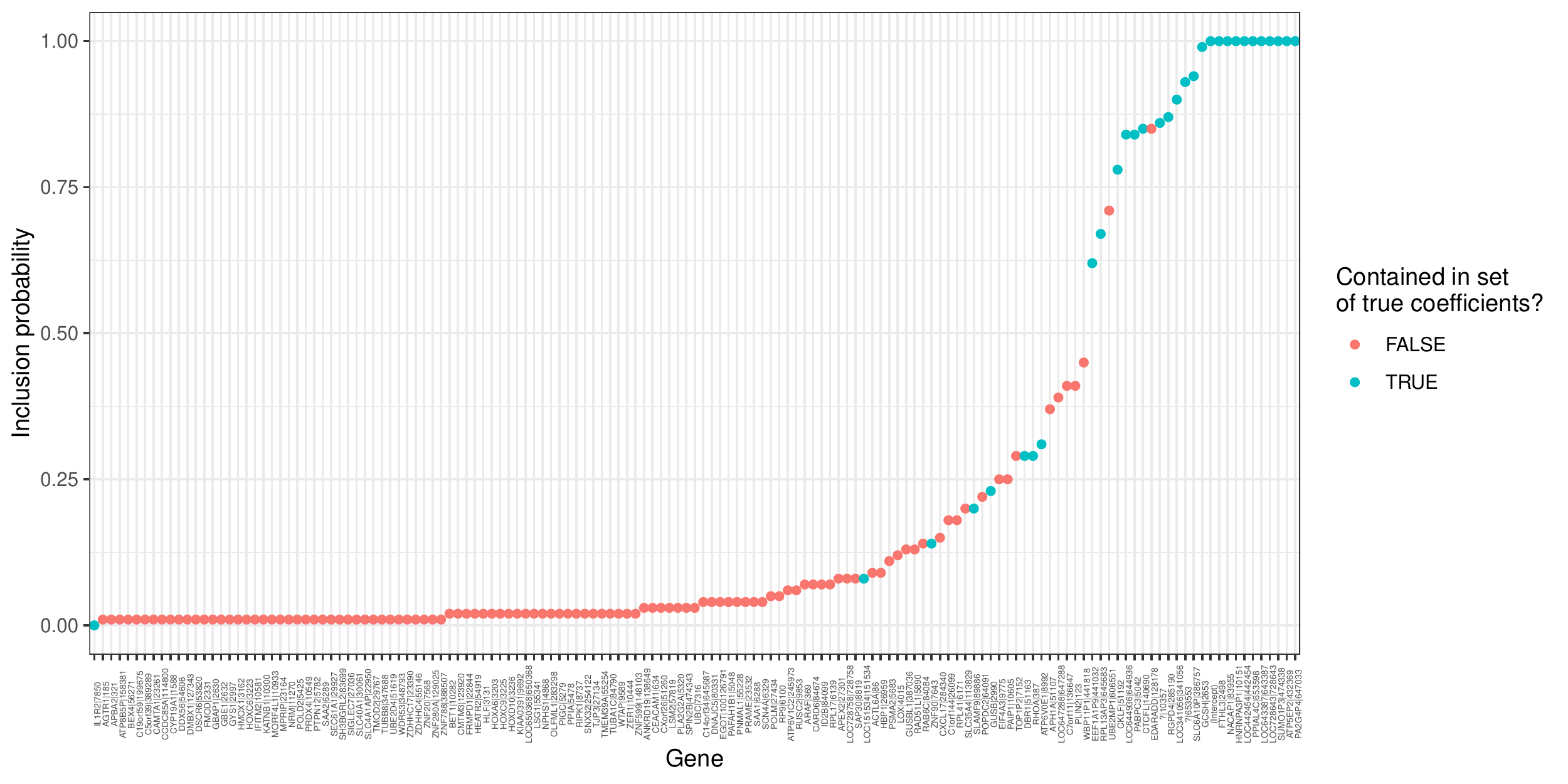}\label{sfig:sfig4b}} \\
\subfloat[SIR]{\includegraphics[width=0.9\textwidth]{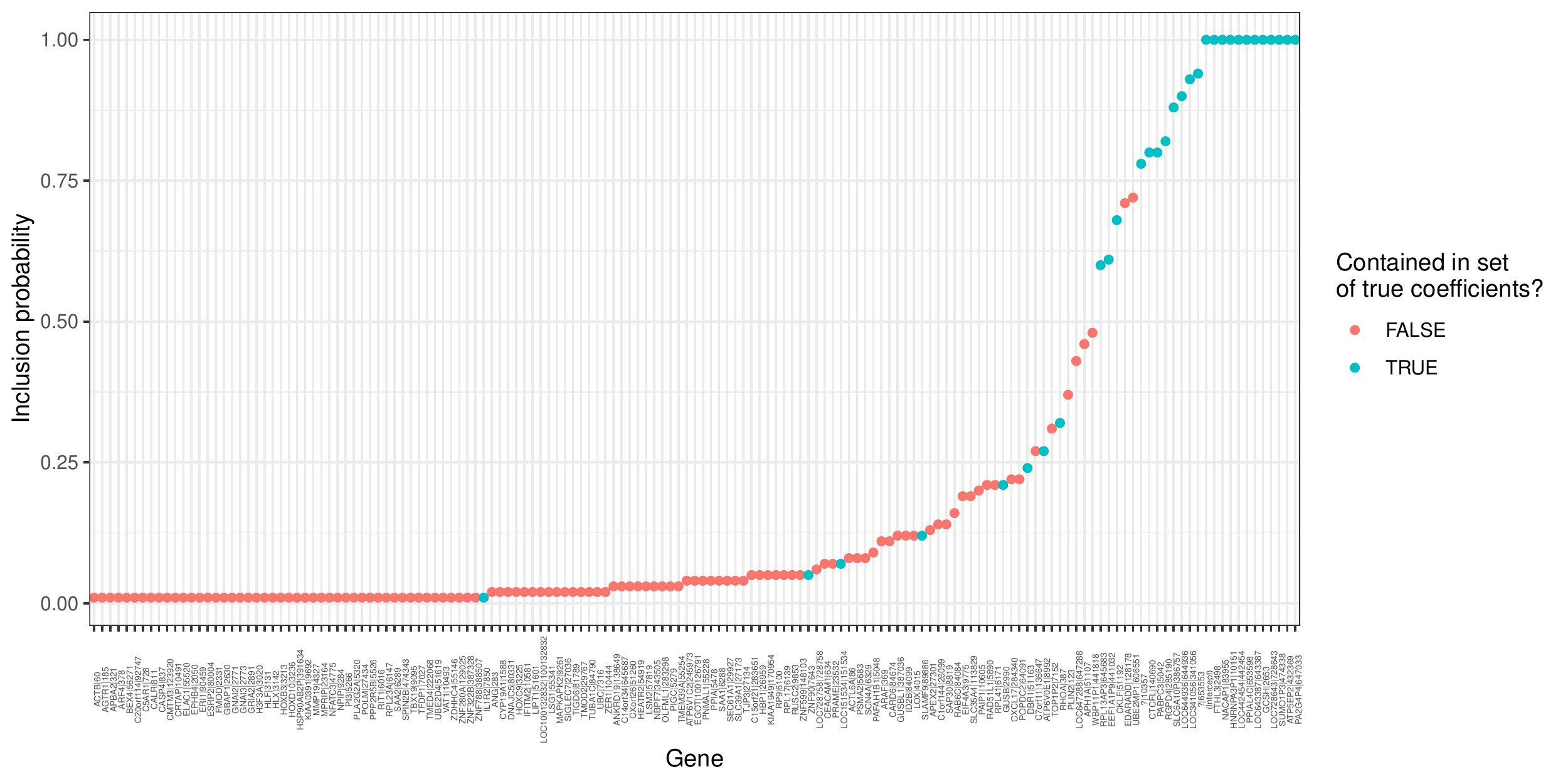}\label{sfig:sfig4c}} \\
\end{center}
\caption{\textbf{Inclusion probability across all non-zero LASSO coefficients across multiple imputation approaches.} Here we report the inclusion probability $P_\text{incl}$ for all non-zero LASSO regression coefficients 
for \textbf{(a)} pseudo-Gibbs (PG), \textbf{(b)} Metropolis-within-Gibbs (MWG) and \textbf{(c)} sampling importance resampling (SIR). Genes in this set that are included in the true LASSO coefficients are highlighted in blue, and those not included in the true discriminating gene set are highlighted in pink.}\label{sfig:sfig4}
\end{figure}

\end{document}